\documentclass{article}

\usepackage[nonatbib,preprint]{neurips_2021}

\usepackage[utf8]{inputenc} 
\usepackage[T1]{fontenc}    
\usepackage[colorlinks]{hyperref} 
\usepackage{url}            
\usepackage{booktabs}       
\usepackage{amsfonts}       
\usepackage{nicefrac}       
\usepackage{microtype}      
\usepackage{xcolor}         

\usepackage{multirow}
\usepackage{graphicx}
\usepackage{amsmath}
\usepackage{threeparttable}
\usepackage{amssymb}
\usepackage{framed,enumitem} 
\usepackage{subfigure}
\usepackage{array}

\newcolumntype{C}{@{\extracolsep{0.7cm}}c@{\extracolsep{5pt}}}%

\newcommand\mmloss{Cross-Modal Code Matching}

\title{Cross-Modal Discrete Representation Learning}

\author{
  Alexander H. Liu \quad SouYoung Jin \quad Cheng-I Jeff Lai \quad Andrew Rouditchenko \\ 
  \\
  \textbf{Aude Oliva \quad James Glass}\\
  \\
  Computer Science and Artificial Intelligence Laboratory\\
  Massachusetts Institute of Technology\\
  Cambridge, MA 02139, USA \\
  \texttt{\{alexhliu, souyoung, clai24, roudi, oliva, glass\}@mit.edu} \\
}

\begin{document}

\maketitle

\begin{abstract}
   Recent advances in representation learning have demonstrated an ability to represent information from different modalities such as video, text, and audio in a single high-level embedding vector.  In this work we present a self-supervised learning framework that is able to learn a representation that captures finer levels of granularity across different modalities such as concepts or events represented by visual objects or spoken words.  Our framework relies on a discretized embedding space created via vector quantization that is shared across different modalities.  Beyond the shared embedding space, we propose a Cross-Modal Code Matching objective that forces the representations from different views (modalities) to have a similar distribution over the discrete embedding space such that cross-modal objects/actions localization can be performed without direct supervision.
   In our experiments we show that the proposed discretized multi-modal fine-grained representation (e.g., pixel/word/frame) can complement high-level summary representations (e.g., video/sentence/waveform) for improved performance on cross-modal retrieval tasks. 
   We also observe that the discretized representation uses individual clusters to represent the same semantic concept across modalities.
\end{abstract}

\section{Introduction}

Toddlers acquire much of their knowledge through grounded learning -- visual concepts can be acquired through language, and language acquisition emerges through visual interaction. 
Inspired by this type of grounded learning, a rich body of representation learning research~\cite{harwath2018jointly,miech2020end,alayrac2020self,monfort2021spoken,luo2021clip4clip} has been exploring the potential to learn from multi-modal data such as video-text, video-audio, and image-audio pairs.
These works typically focus on learning a joint embedding space between different modalities, in which high-level summary representations are extracted as embedding vectors.
These embedding vectors often represent entire video clips, spoken utterances, or sentences as single vectors, and can be useful on tasks such as cross-modal data retrieval, e.g., finding the most similar visual scene according to a spoken language description. 
The predominant approach to learning these embedding vectors is to use modality-independent encoders, and while this has been successful for downstream retrieval tasks, it makes it difficult to compare the activations of the encoders from different modalities.
Further, the space of continuous embedding vectors is unbounded, which makes interpreting the learned representations challenging.

To this end, we propose to jointly learn high-level embedding vector representations with a fine-grained discrete embedding space that is shared across different modalities.
The discrete embedding space enables model interpretability since there are a finite number of embedding vectors which are shared across modalities.
Besides the shared embedding space, we propose a \mmloss~(CMCM) objective that guides the embedding space to capture cross-modal correspondences of concepts, actions, and words.
This not only improves downstream performance on retrieval, but also allows us to better interpret what the model recognized through cross-modal grounded learning.

To verify the effectiveness of our proposed learning framework, we conducted experiments in several cross-modal domains, including video-text, video-audio, and image-audio.
We found consistent improvements over baseline models, verifying that the gain was not restricted to the particular choice of network architecture, input modalities, or dataset.
We also demonstrate the interpretability of the fine-grained discrete representations by showing the cross-modal relations between the embedding vectors and semantic concepts appearing in the input modalities.
Our approach also enables cross-modal concept localization without requiring any labels during training.

\begin{figure*}[t]
\begin{center}
\centerline{\includegraphics[width=\linewidth]{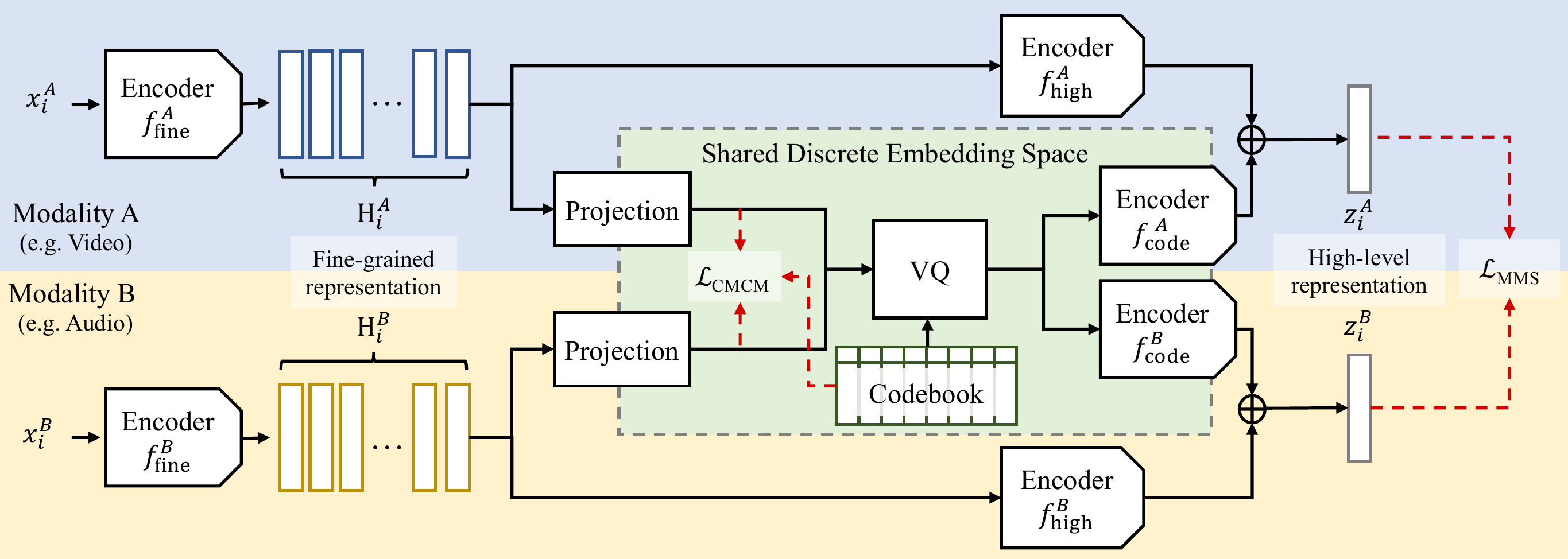}}
\end{center}
\vspace{-20pt}
   \caption{An overview of the proposed framework. The proposed shared discrete embedding space (green region, described in Section~\ref{subsec:vq}) is based on a cross-modal representation learning paradigm (blue and yellow regions, described in Section~\ref{subsec:paradigm}). The proposed \mmloss~$\mathcal{L}_\text{CMCM}$ objective is detailed in Section~\ref{subsec:obj} and Figure~\ref{fig:obj}.   }
\vspace{-10pt}
\label{fig:overview}
\end{figure*}
\vspace{-5pt}
\section{Methodology}
\vspace{-5pt}

Figure~\ref{fig:overview} provides an overview of the proposed framework.
We begin by describing the two-branch cross-modal representation learning paradigm in Section~\ref{subsec:paradigm} (the blue and yellow regions).
Next, we introduce our shared discrete embedding space in Section~\ref{subsec:vq} (the green region).
Finally, in Section~\ref{subsec:obj} and Figure~\ref{fig:obj}, we introduce the \mmloss~objective which guides the model to learn semantically meaningful representations through the shared discrete embedding space.

\vspace{-5pt}
\subsection{Cross-Modal Learning Paradigm}
\vspace{-5pt}
\label{subsec:paradigm}

Given a set of data $\mathcal{X} = {\{(x_i^A,x_i^B)\}}_{i=1}^N$ of size $N$ where each instance $x_i$ is instantiated in different modalities $A$ and $B$ (e.g. video and its corresponding caption), the goal is to derive high-level representative vectors $(z_i^A,z_i^B)$ for each instance $(x_i^A,x_i^B)$ that capture the cross-modal relation measured by a choice of similarity function $S(\cdot,\cdot)$.

For a specific modality $M \in \{A,B\}$, a common first step is to encode raw data $x_i^M$ into a sequence of ``fine-grained'' latent features $H_i^M$ with a modality-specific neural network $f_\text{fine}^M$, i.e. $H_i^M = f_\text{fine}^M(x_i^M)$.
The fine-grained representations $H_i^M$ can express different kinds of raw data, such as video, audio, or sentences, as a sequence of vectors $\{{h_{i,1}^M},...,{h_{i,L}^M}\}$ of length ${L}$.
In the second step, a ``high-level'' representation $z_i^M$ can be derived by summarizing the fine-grained latent features $H_i^M$ with another encoding function $f_\text{high}^M$ that reduces the sequence into a single vector, i.e. $z_i^M = f_\text{high}^M(H_i^M)$.

For example, with modality $A$ being video, raw data $x_i^A$ can be treated as a sequence along time and space and encoded into fine-grained representations $H_i^A=\{{h_{i,l}^A}\}_{l=1}^{L}$ by choosing $f_\text{fine}^A$ to be a Residual Network~\cite{He_2016_CVPR}.
For the second step, a natural choice for $f_\text{high}^A$ to derive the high-level representation $z_i^A$ would be a mean pooling function over the time and spatial axes (arranged along $l$).

With the sets of high-level representations $\{z_i^A\}_{i=1}^N$ and $\{z_j^B\}_{j=1}^N$ from different modalities, we can measure the cross-modal relation between any pair of representations $(z_i^A,z_j^B)$ with some similarity function\footnote{While we used dot product throughtout this work, we also found euclidean distance works well in practice.}$S(\cdot,\cdot)$.
The final step in this paradigm is to adopt an objective function that maximizes the similarity score between ``positive'' pairs (where $i=j$, and thus the true pairs) and minimizes the similarity score between ``negative'' pairs (where $i\neq j$, and thus imposter pairs).

While different objective functions, such as Semi-Hard Negative Mining~\cite{schroff2015facenet} (SHN) and Noise Constrastive Estimation~\cite{pmlr-v9-gutmann10a} (NCE), have been studied in prior work, we focused on the Masked Margin Softmax~\cite{ilharco2019large} (MMS) loss
\begin{equation}
    \mathcal{L}_\text{MMS} = -\frac{1}{N} \sum_{i=1}^{N} \log \frac{e^{S(z_i^A,z_i^B)-M}}{e^{S(z_i^A,z_i^B)-M}+\sum_{j=1}^{N} I_{i\neq j} e^{S(z_i^A,z_j^B)}},
\end{equation}
where the margin $M$ is a hyperparameter to encourage a higher similarity for positive pairs.
The MMS loss $\mathcal{L}_\text{MMS}$ can be seen as an application of the InfoNCE~\cite{oord2018representation} loss with a margin.

The effectiveness of the described cross-modal learning paradigm has been shown by recent works that achieved state-of-the-art results on benchmark datasets in different cross-modal scenarios such as video-text~\cite{luo2021clip4clip}, video-audio~\cite{monfort2021spoken,rouditchenko2020avlnet}, and image-text~\cite{radford2021learning,sanabria2021talk}.

\vspace{-5pt}
\subsection{Shared Discrete Embedding Space}
\vspace{-3pt}
\label{subsec:vq}

While the high-level representations $(z^A_i,z^B_i)$ given by the cross-modal learning paradigm benefit end tasks such as data retrieval, the representations cannot be easily interpreted by humans.
To obtain fine-grained representations that are more interpretable, we introduce a Vector Quantization~\cite{oord2017neural} (VQ) mechanism after obtaining the $H^M_i$ representations.
Formally, with an auxiliary embedding table $E=\{e_1,e_2,...,e_V\}$ of size $V$, which we refer to as the \textit{codebook}, vector quantization is performed on each fine-grained representation ${h_{i,l}^M} \in H^M_i$ of modality $M \in \{A,B\}$ with
\begin{equation}
\label{eq:proj}
    \bar{h}_{i,l}^M = f^M(h_{i,l}^M) + \text{sg}(e_v - f^M(h_{i,l}^M)),
\end{equation}
where $f^M$ is a modality specific projection network to project the input to the shared embedding space, $v = \mathop{\arg\min}_{k\in V} \| h_{i,l}^M - e_k\|_2$, and $\text{sg}(\cdot)$ is the stop-gradient operator proposed in straight-through gradient estimation~\cite{bengio2013estimating} that treats the input as constant during backpropagation.
In other words, each vector ${h_{i,l}^M}$ will be replaced by its nearest neighbor $e_v$, which we refer to as the \textit{codeword}, in the codebook $E$.
The codebook is randomly initialized and updated with the exponential moving average~\cite{oord2017neural} given the fine-grained representations (more details in Section~\ref{sec:ema} of the Appendix).

We trained the shared embedding space jointly with the rest of the framework by modifying the high-level representations $z_i^M$ to include the discretized fine-grained representations as
\begin{equation}
\label{eq:comb}
    z_i^M = f_\text{high}^M(H_i^M) + f_\text{code}^M(\bar{H}^M_i),
\end{equation}
where $f_\text{code}^M$ is, similar to $f_\text{high}^M$, the encoding function for summarizing the sequence of quantized fine-grained representations (e.g., an average pooling function over $l$).
Having such a discrete embedding space allows humans to better interpret the learned embeddings since they are shared across modalities and there are a finite number of them.

\vspace{-5pt}
\subsection{\mmloss}
\vspace{-5pt}
\label{subsec:obj}

\begin{figure*}[t]
\begin{center}
\centerline{\includegraphics[width=\linewidth]{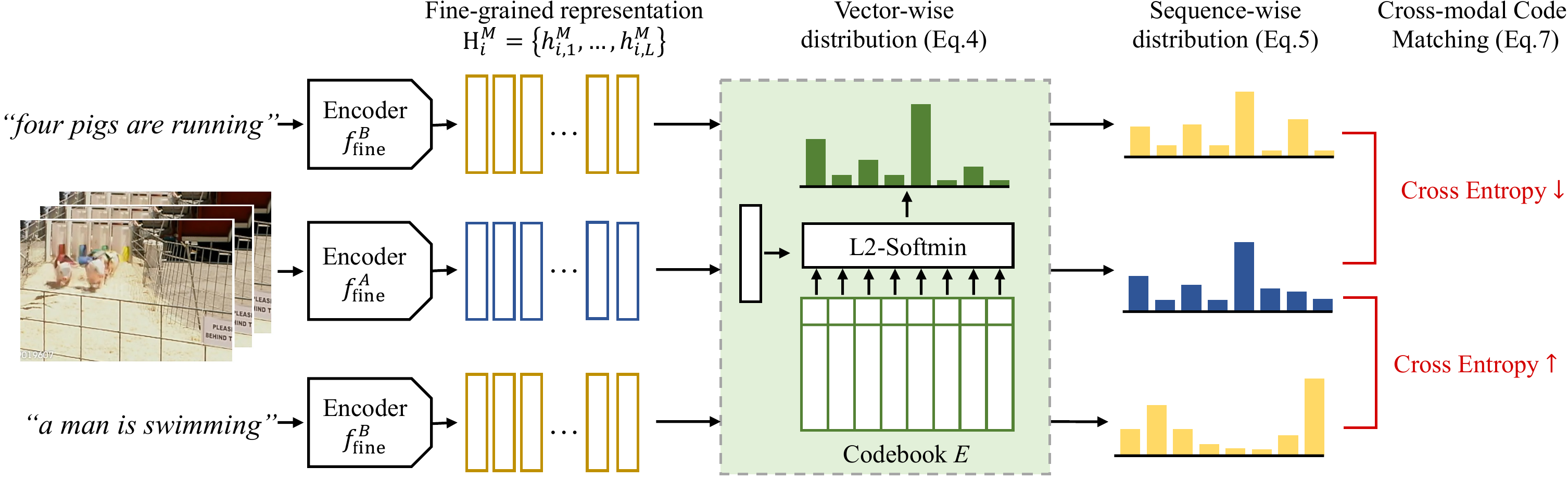}}
\end{center}
\vspace{-20pt}
   \caption{Our proposed Cross-Modal Code Matching objective (described in Section~\ref{subsec:obj}), which encourages the model to use similar codewords for matching cross-modal pairs.
   }
\vspace{-10pt}
\label{fig:obj}
\end{figure*}

Ideally, the codebook should be shared across different modalities since the quantization method is independent to the input modality.
However, as we demonstrate in Section~\ref{sec:no_mcmc} of the Appendix, the model will learn to partition the codebook into modality-specific subspaces due to the significant difference between fine-grained representations from different modalities.
To learn a shared embedding space that is invariant to input modality, we propose the \mmloss~objective which encourages the model to focus more on the semantic aspect of the input, as illustrated in Figure~\ref{fig:obj}.

For each vector $h_{i,l}^M$ in the fine-grained representation sequence $H_i^M$ encoded from an instance $x_i^M$ of modality $M$, we first define the probability of $h_{i,l}^M$ belonging to the codeword $e_v$ as the Softmin function of their Euclidean distance, that is
\begin{equation}
    \label{eq:prob}
    P(e_v|h_{i,l}^M) = \frac{\exp(- \| f^M(h_{i,l}^M) - e_v\|_2)}{\sum_{k \in V} \exp(- \| f^M(h_{i,l}^M) - e_k\|_2)}.
\end{equation}
Note that this definition assigns higher a probability to codewords that are closer to the fine-grained representation, where the closest codeword is used to perform vector quantization.
We can then define the sequence-level probability distribution over the codebook as the average of the fine-grained distribution, that is
\begin{equation}
    P(e_v|H_i^M) = \frac{1}{L} \sum_{l} P(e_v|h_{i,l}^M).
\end{equation}
This distribution is essentially the normalized frequency of codeword usage for a given sequence of fine-grained representations.
Next, for a pair of cross-modal data $(x_i^A,x_j^B)$, we define their \textit{code similarity} as the negative symmetric cross entropy of probability distribution over the codebook
\begin{equation}
\begin{aligned}
\label{eq:sce}
    S_\text{code}(x_i^A,x_j^B) &= \sum_v P(e_v|H_i^A) \log P(e_v|H_j^B)
    + \sum_v P(e_v|H_j^B) \log P(e_v|H_i^A), \\
\end{aligned}
\end{equation}

Finally, we propose the \textit{Cross-Modal Code Matching (CMCM)} objective using code similarity as
\begin{equation}
    \mathcal{L}_\text{CMCM} = -\frac{1}{N} \sum_{i=1}^{N} \log \frac{e^{S_\text{code}(x_i^A,x_i^B)}}{e^{S_\text{code}(x_i^A,x_i^B)}+\sum_{j\neq i} e^{S_\text{code}{(x_i^A,x_j^B)}}}.
\end{equation}

Intuitively, the proposed objective encourages the model to represent the input $(x_i^A,x_j^B)$ with similar codewords for positive pairs ($i=j$) and non-matching codewords for negative pairs ($i\neq j$).
As a consequence, each codeword is expected to be a modality invariant representation of a more fine-grained concept, action, or word that can be discovered from cross-modal data.
For example, a codeword could correspond to both the visual scene of a man juggling, and also the spoken word ``juggling,'' as we demonstrate in our experimental results in Table~\ref{tb:cooccur} and Figure~\ref{fig:demo}.

The full objective of our proposed cross-modal representation learning framework is the combination of objectives at different levels
\begin{equation}
    \mathcal{L} = \mathcal{L}_\text{MMS} + \alpha \mathcal{L}_\text{CMCM},
\end{equation}
where $\alpha$ controls the weight between the two terms. 
Empirically, we found $\alpha=0.1$ worked well across different settings.

\vspace{-5pt}
\section{Related work}
\vspace{-5pt}
\label{sec:related}

\textbf{Methods fitting into the cross-modal learning paradigm.}
As described in Section~\ref{subsec:paradigm}, many of the existing methods for cross-modal learning fit into the paradigm where encoders are modality-independent.
This paradigm has been shown to be effective by achieving state-of-the-art retrieval performance on benchmark datasets with the modality pairs that we considered in this work: video-text~\cite{bain2021frozen,luo2021clip4clip}, video-audio~\cite{monfort2021spoken,rouditchenko2020avlnet}, and image-audio~\cite{harwath2018jointly,Harwath2020Learning}.
While these prior works relied on different pre-training datasets, model architectures, and objective functions, they all leverage modality-independent encoders.
One of the most important features of this paradigm is the fixed inference time for retrieval.
Since the encoders are modality-independent, embedding vectors for samples in a given modality can be computed without using any samples from the other modality.
Thus retrieval only involves computing the dot product between embedding vectors from two different modalities.
As a consequence, these models are more flexible for large-scale retrieval, and the embedding vectors from each modality can be used independently for other downstream tasks.

\textbf{Other cross-modal learning frameworks.}
\label{subsec:other_model}
In contrast to the aforementioned works, some methods leverage cross-modal relations within the encoders instead of using modality-independent encoders.
This has been done with both cross-modal encoders~\cite{lei2021less,luo2021clip4clip} and cross-modal attention mechanisms~\cite{miech2018learning,liu2019use,Liu_2019_CVPR,gabeur2020multi}.
However, the cross-modal interactions increase the complexity for retrieval since every instance of a specific modality must be used as input with every instance of another modality to obtain the embedding vectors.
With $m$ and $n$ samples in the modalities respectively, this increases the complexity from the modality-independent approach from $\mathcal{O}(m+n)$ to $\mathcal{O}(mn)$.
Further, it also makes analysis of the embedding vectors from any individual modality challenging and inhibits single-modality downstream tasks.
Our proposed framework builds on the modality-independent approach to enable light-weight retrieval, but it also enables cross-modal interaction through our proposed codebook and \mmloss~objective.

\textbf{Uncovering semantic-level correspondences.}
Image-audio models have been shown to discover spoken words and visual objects without supervision through retrieval tasks~\cite{synnaeve2014learning,harwath2015deep,harwath2017unsupervised,kamper2018semantic}, and the audio embedding vectors have been shown to cluster into word-like speech units~\cite{harwath2017learning,wang2019multimodal,Harwath2020Learning}.
Some work has studied the ability of video-audio models to relate spoken words to visual objects and actions in videos~\cite{boggust2019grounding,rouditchenko2020avlnet}.
However, none of these models incorporated a shared embedding space that enabled modality-invariant representations.
VQ units have been used in the audio encoder of an image-audio model~\cite{Harwath2020Learning}, which allowed it to capture the hierarchical structure of spoken language.
While our proposed framework is similar in that it also discretizes the audio sequence with VQ units, our work differs significantly by capturing the cross-modal interactions between visual and audio inputs in the shared embedding space rather than solely capturing the tree structure of speech.
Further, besides image-audio data, our proposed framework can handle video-audio and video-text data.
Finally, modality invariant audio-visual representations have been explored using variational autoencoders~\cite{leidal2017learning,zhu2021learning}, while we propose modality invariant representations for different cross-modal domains using a shared discrete embedding space.

\vspace{-5pt}
\section{Experiments}
\vspace{-5pt}
\label{sec:exp}

\subsection{Setup}
\label{subsec:setup}

To demonstrate the generalizability of the proposed method, we tested our framework on different cross-modal datasets and baseline models that fit into the cross-modal learning paradigm.
All setups are listed below and summarized in Table~\ref{tb:setup} of the Appendix.
For training the proposed model, we randomly initialized all the modules related to the discrete shared embedding space and trained them jointly with the rest of the framework (see Figure~\ref{fig:overview}).
Unless otherwise specified, (1) we ``warm-started'' our proposed framework by initializing it with the modality-specific encoders (namely, $f^M_\text{fine}$ and $f^M_\text{high}$) from the baseline models;
(2) both the projection network $f^M$ and the encoder network $f_\text{code}^M$ are single linear layers; (3) the codebook size is set to 1024.
Please refer to Section~\ref{sec:detail} in the Appendix for more implementation details and computational costs.

\textbf{Video-Text: MSR-VTT}~\cite{xu2016msr} contains 10k video clips with length varying from 10 to 32 seconds.
While each video is provided with 20 related captions for training, we followed the evaluation protocol from previous works~\cite{luo2021clip4clip,gabeur2020multi,yu2018joint} to use the \texttt{training-9k}~/~\texttt{test 1k-A} splits for training and testing respectively.
CLIP4Clip~\cite{luo2021clip4clip}, the current state-of-the-art on MSR-VTT, is selected as the baseline model.
Following the cross-modal learning paradigm described in Section~\ref{subsec:paradigm}, CLIP4Clip is composed of a pair of encoders: a Visual Transformer~\cite{dosovitskiy2020image} and a Text Transformer~\cite{vaswani2017attention}.
Both encoders are initialized from the CLIP model~\cite{radford2021learning}, which is pre-trained on the text-image dataset WIT~\cite{radford2021learning} and optimized in the end-to-end manner from pixel/text input.
For training the proposed framework on top of CLIP4Clip, we freeze the transformers from CLIP4Clip and update only the modules related to the discrete shared embedding space.
Both the projection network $f^M$ (see Eq.~\ref{eq:proj}) and the encoder network $f_\text{code}^M$ (see Eq.~\ref{eq:comb}) are 4D-Convolutions for video with a depth of 3 and BiLSTMs for text, also with a depth of 3.
While CLIP4Clip provided different options for the high-level visual encoder $f_\text{high}^M$, we adopted the vanilla mean-pooling model.
Following CLIP4Clip, the shared embedding space has a dimension of 512.

\textbf{Video-Audio: S-MiT}~\cite{monfort2021spoken} contains over 500k pairs of 3-second video and corresponding spoken audio captions averaging 8 seconds.
We followed the official protocol to train on the training set of 500k pairs, use the validation set of 10k pairs for development and analysis, and report the retrieval result on a 1k search space over 5 runs randomly sampled from a held-out test set.
We selected the same baseline model used on the dataset~\cite{monfort2021spoken}, which contains a visual encoder composed of a ResNet-152 pre-trained on ImageNet~\cite{deng2009imagenet} and TSM ResNet-50~\cite{lin2019tsm} pre-trained on M-MiT~\cite{monfort2019multi}.
The audio encoder is a randomly initialized 1D-ResNet~\cite{harwath2020jointly} designed specifically for spectrograms.
The shared embedding space has the dimension of 4096, matching the encoders in the baseline model.

\textbf{Image-Audio: Places}~\cite{harwath2017unsupervised} contains over 400k pairs of images from the Places 205 dataset~\cite{zhou2014learning} and corresponding spoken audio captions averaging 10 seconds.
We followed the previous works~\cite{harwath2018jointly,harwath2020jointly,Harwath2020Learning} to use the training set of 400k pairs and report results on the validation set of 1k pairs.
We select ResDAVEnet~\cite{harwath2020jointly} as the baseline model where the visual encoder is a ResNet-50 pre-trained on ImageNet~\cite{deng2009imagenet} and the audio encoder is a randomly initialized 1D-ResNet~\cite{harwath2020jointly} designed specifically for spectrograms.
The shared embedding space has the dimension of 1024.

\begin{table*}
\centering
\begingroup
\setlength{\tabcolsep}{5pt}
\renewcommand{\arraystretch}{1.2}
\begin{threeparttable}
  \caption{\normalsize{Cross-Modal retrieval results on MSR-VTT, S-MiT, and Places.}}
  \label{tb:retrieve}
  \centering
  \small
  \begin{tabular}{l  c c c c  c c c c  }
    \toprule
     Modality \textit{A}-\textit{B} / Dataset & \multicolumn{4}{c}{ Visual Retrieval }  & \multicolumn{4}{c}{ Language Retrieval } \\
     \multirow{2}{*}{~~~~~~Method} & \multicolumn{4}{c}{ (\textit{B} $\rightarrow$ \textit{A}) }  & \multicolumn{4}{c}{ (\textit{A} $\rightarrow$ \textit{B}) } \\
     \cmidrule(lr){2-5}\cmidrule(lr){6-9}
     & R@1 $\uparrow$ & R@5 $\uparrow$ & R@10 $\uparrow$ & MnR $\downarrow$ & R@1 $\uparrow$ & R@5 $\uparrow$ & R@10 $\uparrow$ & MnR $\downarrow$  \\
    \hline\hline
    \multicolumn{9}{l}{Video-Text / MSR-VTT~\cite{xu2016msr}}\\ \hline
    ~~Frozen-in-Time~\cite{bain2021frozen} & 31.0 & 59.5 & 70.5 & - & - & - & - & - \\
    ~~CLIP4Clip-meanP~\cite{luo2021clip4clip} & 43.1 & 70.4 & 80.8 & 16.2 & - & - & - & - \\
    ~~CLIP4Clip-tightT~\cite{luo2021clip4clip} & 40.2 & 71.5 & 80.5 & \textbf{13.4} & - & - & - & - \\
    ~~Our Baseline{\textdagger} & 42.6 & 71.2 & 80.8 & 15.5 & \textbf{43.0} & 70.9 & 80.9 & 12.5 \\
    ~~\textbf{Proposed} & \textbf{43.4} & \textbf{72.3} & \textbf{81.2} & 14.8 & 42.5 & \textbf{71.2} & \textbf{81.1} & \textbf{12.0} \\
    \hline\hline
    \multicolumn{9}{l}{Video-Audio / S-MiT~\cite{monfort2021spoken}}\\ \hline
    ~~S-MiT~\cite{monfort2021spoken} & 32.1 & 58.9 & 68.6 & -  & 32.3 & 57.9 & 68.1 & - \\
    ~~Our Baseline{\textdagger} & 30.2 & 57.3 & 68.5 &41.9  & 29.7 & 57.2 & 68.7 &  28.5 \\
    ~~\textbf{Proposed} & \textbf{34.3} & \textbf{61.3} & \textbf{72.0} & \textbf{33.5} & \textbf{34.0} & \textbf{61.6} & \textbf{71.7}  &  \textbf{22.5}\\
    \hline\hline
    \multicolumn{9}{l}{Image-Audio / Places~\cite{harwath2017unsupervised}} \\ \hline 
    ~~ResDAVEnet~\cite{harwath2020jointly}* & 30.9 & 63.6 & 74.2 & 20.2 & 26.4 & 58.5 & 71.2 & 21.6 \\
    ~~ResDAVEnet-VQ~\cite{Harwath2020Learning}* & 34.9 & 70.2 & 79.4 & 15.0 & 32.7 & 65.6 & 77.0 & 18.0 \\
    ~~Our Baseline{\textdagger} & 43.8 & 74.1 & 82.4 & 15.8 & 40.4 & 73.3 & 82.5 & 10.9 \\
    ~~\textbf{Proposed} & \textbf{46.5} & \textbf{77.4} & \textbf{85.8} & \textbf{13.7} & \textbf{45.4} & \textbf{77.7} & \textbf{85.9}  &  \textbf{8.9}\\
    \bottomrule
  \end{tabular}
\begin{tablenotes}
\item{\textdagger} \small{\text{Existing model reproduced with $\mathcal{L}_\text{MMS}$ for fair comparison, see Table~\ref{tb:setup} in the Appendix for more detail.}}
\item{*} \small{\text{Results obtained by running the official code and pre-trained models, see Appendix for more details.}}
\end{tablenotes}
\end{threeparttable}
\vspace{-10pt}
\endgroup
\end{table*}

\subsection{Cross-Modal Retrieval}
\label{subsec:retreival}

Data retrieval is one of the most common evaluations for cross-modal representation learning.
For example, in video retrieval with input query text, videos in the search space will be ranked by the similarity between the representation of each video and the query.
We report the standard retrieval metrics recall at rank K (R@K) and median rank (MdR) in Table~\ref{tb:retrieve}.
We show the performance on both visual retrieval, where input language queries are used to retrieve videos or images, and language retrieval, where input visual queries are used to retrieve spoken or text captions.

\textbf{Video-Text Retrieval.} On the benchmark MSR-VTT dataset, we compared our proposed method against recent works achieving state-of-the-art~\cite{bain2021frozen,liu2021hit,luo2021clip4clip} and provide a full comparison against more prior work~\cite{liu2019use,rouditchenko2020avlnet,gabeur2020multi,patrick2020support,dzabraev2021mdmmt,croitoru2021teachtext} in Section~\ref{sec:full_msrvtt} of the Appendix.
Frozen-in-Time~\cite{bain2021frozen} and CLIP4Clip~\cite{luo2021clip4clip} are similar methods that employ a Visual Transformer~\cite{dosovitskiy2020image} to encode video as sequence of images.
The key differences between them is the choice of summarizing function (i.e. $f^M_\text{high}$) for video and the pre-training procedure.
We also note that the CLIP4Clip with tight transformer encoder~\cite{luo2021clip4clip} (CLIP4Clip-tightT) relied on cross-modal reference via self-attention encoders to derive representations, which has a higher time complexity as mentioned in Section~\ref{sec:related}.
With the shared codebook and \mmloss~objection, our proposed framework also enables cross-modal reference and gives an improvement over the baseline model without increasing the time complexity.

\textbf{Video-Audio Retrieval.} Video-Audio retrieval on S-MiT~\cite{monfort2021spoken} is a challenging task since videos are paired with raw speech audio, which is untranscribed, unsegmented, and can contain background noise and speaker variation.
However, our proposed framework that leverages cross-modal connections between visual actions and spoken words is able to improve the baseline model by a margin.
We further analyze our framework's ability to relate visual actions and spoken words in Section~\ref{subsec:codeword_exp}.

\textbf{Image-Audio Retrieval.} 
We compare our proposed method against the recent models~\cite{harwath2020jointly,Harwath2020Learning} achieving state-of-the-art and provide a full comparison to previous methods~\cite{harwath2018jointly,harwath2017learning,harwath2017unsupervised} in Section~\ref{sec:full_places} of the Appendix.
Comparing the baseline model, ResDAVEnet~\cite{harwath2020jointly}, and the current state-of-the-art ResDAVEnet-VQ~\cite{Harwath2020Learning}, the latter model introduces VQ units into the audio encoder, allowing it to model the hierarchical structure of speech and achieve better retrieval results.
With our framework, we introduce our shared VQ embedding space into the ResDAVEnet model to capture cross-modal interactions.
This improves the performance over both ResDAVEnet and ResDAVEnet-VQ.

Overall, our proposed method enables consistent improvements regardless of the data modalities and baseline architectures, demonstrating its effectiveness and generalizability.

\subsection{Discrete Representation Analysis}
\label{subsec:codeword_exp}
\begin{figure*}[t]
\begin{center}
\centerline{\includegraphics[width=\linewidth]{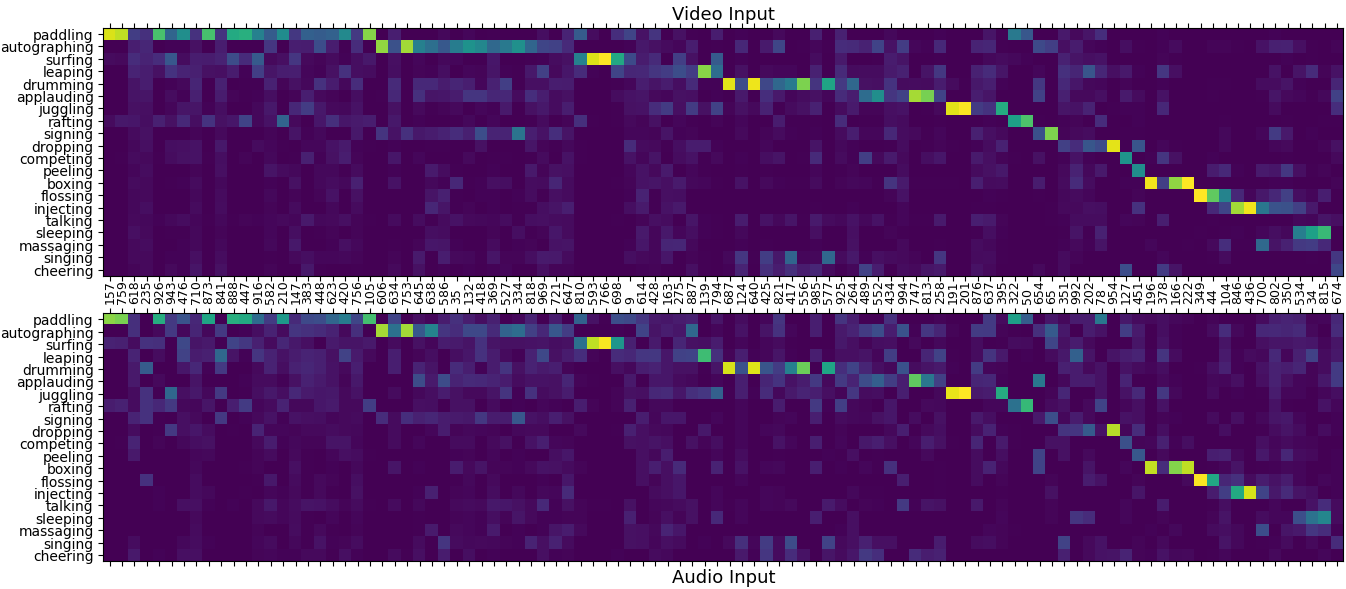}}
\end{center}
\vspace{-20pt}
\caption{Conditional probability matrix illustrating $P(\text{action}|\text{codeword})$ on the S-MiT development set. Y-axis is action label, showing only the top 20 most frequent labels for simplicity. X-axis is the indices of the top 100 most frequent codewords. }
\vspace{-10pt}
\label{fig:cor}
\end{figure*}

One of the important motivations of introducing the discrete cross-modal embedding space is better model interpretability.
In this section, we take a closer look into the codewords learned through our proposed framework.
For the evaluation, we chose the video-audio setup on S-MiT~\cite{monfort2021spoken}.
We used video-audio pairs from the development set, where each pair is labeled with an action out of 332 categories.
Note that we only used labels for analysis, labels are never used for training.

\textbf{Conditional Probability of Action Labels Given Codeword.}
First, we compute the conditional probability distributions of action labels given the codewords over the video inputs.
Each video input is fixed-length and represented by 27 codewords (3 frames each represented by 3$\times$3 codewords), and we labeled all these codewords with the video's action label.
By accumulating codeword labels through the whole development set, we can compute the conditional probability of each action given any codeword, i.e. $P(\text{action}|\text{codeword})$.
Results are visualized in the upper part of Figure~\ref{fig:cor}.
Similarly, we computed the conditional probabilities based on the audio input where each utterance is represented by up to 32 codewords depending on the utterance length.
We selected the most frequent codewords used by the video inputs and plot the conditional probabilities based on the audio input in the lower part of Figure~\ref{fig:cor}.
We can observe that both matrices have similar patterns, i.e., when a codeword is activated, there is a high chance of a specific action appearing in the input regardless if it is video or audio.
This suggests that our model is able to learn cross-modal representations for actions grounded by either visual or spoken language input. 
The codewords are not only modality invariant, but more importantly, they also capture the semantic relations of the labels.
e.g., codewords with the highest chance to represent ``autographing'' typically have the second highest chance of representing ``signing''; codewords for ``surfing'' are less likely to represent other actions as all of them are very different from ``surfing''.
We also note that without the \mmloss~objective, semantically related video and audio inputs no longer use the same codewords, which we illustrate in Section~\ref{sec:no_mcmc} of the Appendix.

\begin{table*}
\centering
\begingroup
\setlength{\tabcolsep}{5pt}
\renewcommand{\arraystretch}{1.2}
  \caption{Correspondence between codewords, visual actions, and spoken words. Ranking is based on the precision (Prc.) of the top hypothesis of the visual action label. Occurrence (Occ.) indicates the number of times the codeword was activated throughout the development set. Around 750 codewords were activated on the development set. An extended table is available in Section~\ref{sec:more_demo} of the Appendix.}
  \label{tb:cooccur}
  \centering
  \small
  \begin{tabular}{c  c  c c c c c  c c c c }
    \toprule
     \multirow{3}{*}{Rank} &\multirow{3}{*}{Code}  & \multirow{3}{*}{Occ.} &  \multicolumn{4}{c}{ Visual Action }  & \multicolumn{4}{c}{ Spoken word } \\
     \cmidrule(lr){4-7}\cmidrule(lr){8-11}
      & & & \multicolumn{2}{c}{ Top Hypothesis } & \multicolumn{2}{c}{ Second Hypothesis }  &  \multicolumn{2}{c}{ Top Hypothesis } & \multicolumn{2}{c}{ Second Hypothesis } \\
     & & & Label & Prc. & Label & Prc.  & Word & F1 &  Word & F1  \\
    \hline\hline
1 & 201 & 147 & juggling & 97.5 & kicking & 1.2 & juggling & 36.7 & juggles & 8.3\\
2 & 349 & 112 & flossing & 96.0 & licking & 0.7 & floss & 15.8 & flossing & 14.0\\
3 & 145 & 49 & surfing & 95.6 & snowing & 2.9 & surfboard & 23.7 & waves & 7.3\\
4 & 29 & 64 & tattooing & 94.6 & injecting & 2.2 & tattoo & 15.8 & tattooed & 4.2\\
5 & 233 & 25 & ironing & 93.8 & hammering & 6.2 & ironing & 20.5 & iron & 4.7\\
\multicolumn{11}{c}{...}\\
32 & 500 & 89 & dialing & 60.0 & texting & 10.0 & dialing & 13.8 & phone & 9.8\\
33 & 536 & 28 & cheering & 60.0 & shouting & 10.0 & cheerleaders & 26.8 & cheerleading & 10.3\\
34 & 50 & 203 & rafting & 58.6 & paddling & 25.7 & rafting & 16.7 & raft & 8.5\\
35 & 664 & 78 & dunking & 58.0 & leaping & 9.1 & basketball & 11.0 & dunking & 5.2\\
\multicolumn{11}{c}{...}\\
742 & 733 & 188 & discussing & 6.5 & applauding & 4.6 & men & 7.3 & two & 6.4\\
743 & 542 & 58 & baking & 6.5 & peeling & 5.2 & cupcake & 9.2 & peanut & 6.2\\

    \bottomrule
  \end{tabular}
\endgroup
\end{table*}
\begin{figure*}[t]
\begin{center}
\centerline{\includegraphics[width=1.0\linewidth]{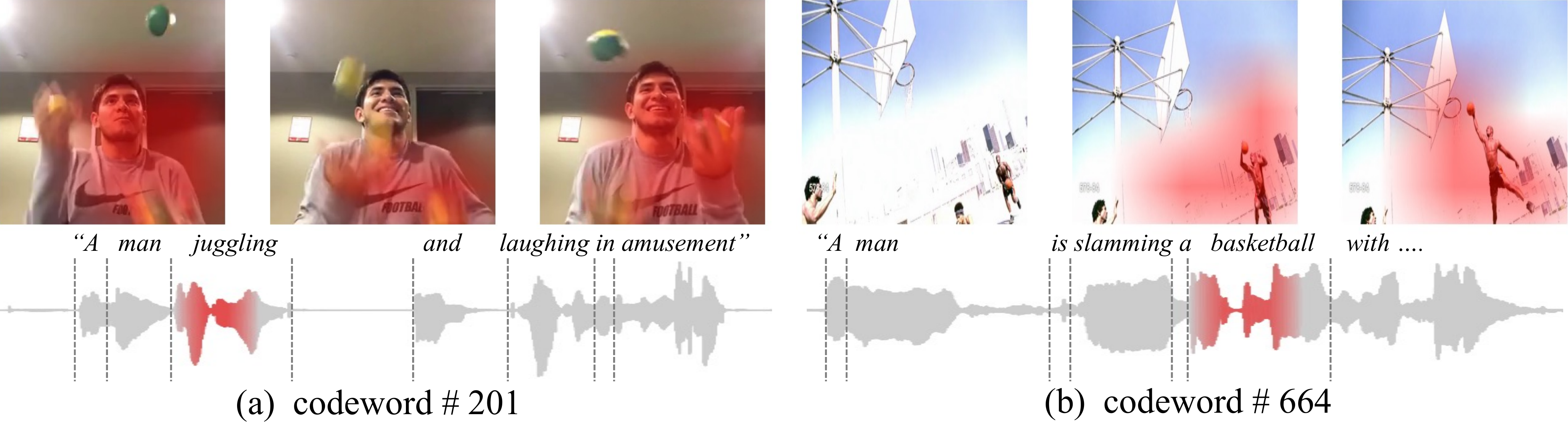}}
\end{center}
\vspace{-20pt}
\caption{Codeword cross-modal localization. Input regions that are encoded by the codeword (selected from Table~\ref{tb:cooccur}) are highlighted in red.}
\vspace{-10pt}
\label{fig:demo}
\end{figure*}

\textbf{Cross-Modal Correspondences.} Next, we analyze the connections captured by the codewords between action labels and spoken words.
With the same label accumulation method described previously, we compute the precision of action prediction with codewords (i.e. $\frac{\text{code-action co-occurrence}}{\text{code occurrence}}$).
For the audio, we used word-level transcriptions (from Google's speech-to-text API) to assign a spoken word to each codeword when it is activated by the input utterance.
This results in a hypothesis set including around 7k words for each codeword, and we listed the top 2 hypotheses for each codeword with the highest F1 score (instead of precision to avoid domination of high-frequency words).
Results are listed in Table~\ref{tb:cooccur}.
For the codewords that have the highest precision on predicting the action label, we found the top hypotheses for spoken words are often the action label itself.
E.g., the codeword (rank 1st) for the visual action ``juggling'' maps to the spoken word ``juggling'' perfectly.
As precision on visual action prediction decreases, we observed fewer perfect mappings, but the spoken word hypotheses remained semantically related to the visual action hypotheses.
E.g., the codeword (rank 35th) for the visual action ``dunking'' with lower precision now maps to the spoken word ``basketball.''
Surprisingly, even the codewords with the lowest precision capture relationships between visual actions and spoken words to some extent.
E.g., codeword (rank 743th) that is most related to the action ``baking'' has the top and second word hypotheses ``cupcake'' and ``peanut.''

\textbf{Codeword Localization.} Finally, to visualize the relation between codewords and the input data, we localize the segments of both the video and audio input that are assigned to certain codewords.
This is possible because quantization in our shared embedding space is done at the fine-grained level, so that the time and spatial axes are preserved.
Examples are shown in Figure~\ref{fig:demo}, where the regions assigned to the given code are highlighted.
Interestingly, we see the codewords being aligned to both the visual actions and the corresponding spoken words.
This supports our claim of having a more interpretable representation at the fine-grained level.

\subsection{Ablation Study}
\label{subsec:ablation}

\begin{table*}
\centering
\begingroup
\setlength{\tabcolsep}{5pt}
\renewcommand{\arraystretch}{1.2}
\small
  \caption{Ablation study performed on Places~\cite{harwath2017unsupervised}, scores are averaged over audio and image retrieval.}
  \label{tb:ablation}
  \centering
  \begin{tabular}{c l  c c c c  }
    \toprule
     \multicolumn{2}{l}{\multirow{2}{*}{~~~~~~Method}} & \multicolumn{4}{c}{ Averaged 2-way Retrieval }   \\
     & & R@1 $\uparrow$ & R@5 $\uparrow$ & R@10 $\uparrow$ & MnR $\downarrow$ \\
    \hline\hline
    ~~(a)& Proposed & 46.0 & 77.6 & 85.9 & 11.3 \\ \hline
    ~~(b)& codebook size = 512 & 46.2 & 77.4 & 85.2 & 11.5 \\
    ~~(c)& codebook size = 2048  & 46.1 & 76.6 & 84.7 & 12.1 \\
    ~~(d)& $\alpha=1.0$ & 45.6 & 76.6 & 85.5 & 11.6 \\
    ~~(e)& $\alpha=0.0$ (w/o Cross-Modal Code Matching) & 45.2 & 75.5 & 84.2 & 12.8 \\
    ~~(f)& w/o VQ \& w/o Cross-Modal Code Matching & 45.7 & 75.9 & 84.7 & 12.6 \\
    ~~(g)& w/o warm-start & 41.6 & 73.4 & 82.5 & 16.0 \\
    ~~(h)& w/o continuous representation ($f_\text{high}^M(H_i^M)$) & 29.0 & 63.0 & 74.7 & 19.4 \\
    ~~(i)& Our Baseline & 42.1 & 73.7 & 82.5 & 13.4 \\
    
    \bottomrule
  \end{tabular}
 \vspace{-10pt}
\endgroup
\end{table*}
To justify our framework design and choice of hyperparameters, we conducted an ablation study on the image-audio setting and report the results in Table~\ref{tb:ablation}.

\textbf{Impact of the shared embedding space.} 
For the codebook size, 1024 codewords worked well across different datasets.
Halving and doubling the number of codewords (row(b) \& (c)) both decreased the performance slightly.
For the weight $\alpha$ of the proposed \mmloss~objective, we found that values in the range $(0,1]$ generally work while $0.1$ works the best (row(a) v.s. row(d)).
Removing the proposed \mmloss~objective (setting $\alpha=0$, row(e)), however, hurts the performance.
Furthermore, without the objective, the codebook no longer captures cross-modal correspondences, as illustrated in Section~\ref{sec:no_mcmc} of the Appendix.
We also observed that disabling the VQ layer together with the \mmloss~objective slightly recovers performance (row(f) v.s. row(e)).
All of these observations serve as evidence that the proposed discrete embedding space is most beneficial to the retrieval task with the guidance from the \mmloss~objective.

\textbf{Importance of baseline models in the cross-modal learning paradigm.} As mentioned in Section~\ref{subsec:setup}, the discrete shared embedding space is learned with ``warm-starting'' from a baseline model.
We note that warm-starting is important for getting more refined representations that yield better retrieval results (row(a) v.s. row(g)).
Without warm-starting, our framework can only perform similar to the baseline (row(g) v.s. row(i)).
This finding aligns with previous work~\cite{Harwath2020Learning} that used VQ layers in the audio encoder and used warm-starting to learn acoustic units.
Moreover, removing the continuous representations (row(h)) originally used in the cross-modal learning paradigm and using only the codeword representations significantly decreases performance.
This exposes the trade-off between interpretability and end-task performance by imposing a discrete embedding space.
Hence, we choose to integrate both discrete and continuous embedding space for retrieval as in Eq.~\ref{eq:comb}.

\section{Conclusion}
\label{sec:conclusion}
In this paper, we proposed a framework for cross-modal representation learning with a discrete embedding space that is shared amongst different modalities and enables model interpretability.
We also propose a \mmloss~objective that encourages models to represent cross-model semantic concepts in the embedding space.
Combining our discrete embedding space and objective with existing cross-modal representation learning models improves retrieval performance on video-text, video-audio, and image-audio datasets. 
We also analyze the shared embedding space and find that semantically related video and audio inputs tend to use the same codewords.

\textbf{Limitations.}
As described in Section~\ref{subsec:paradigm}, the present work relies on the existing cross-modal learning paradigm with modality-independent encoders.
Although our proposed method is shown to be effective, further work is required to generalize our method to other cross-modal learning frameworks which are more computationally complex.
In addition, recent work~\cite{alayrac2020self,rouditchenko2020avlnet} demonstrates the benefits of learning from three or more modalities.
While the proposed method can theoretically be extended to more modalities, further work is required to handle practical bottlenecks such as the growth in time complexity for the \mmloss~objective with respect to the number of modalities. 

\textbf{Potential Negative Societal Impacts.} 
Our proposed method is self-supervised and does not use any labels during training.
It will therefore learn any biases present in the data.
However, we expect that our discrete embedding space will help improve interpretability and could help humans discover the biases present in training data before deploying models to any real-world applications.

\begin{ack}
This research was supported by the MIT-IBM Watson AI Lab. We thank Dave Harwath for helpful discussions.
\end{ack}


{\small
\bibliographystyle{ieee}
\bibliography{main}
}


\newpage
\appendix

\section*{Appendix}

\section{Codebook Update Policy}
\label{sec:ema}
The codebook with $d$-dimensional codewords is initialized with
\begin{equation}
\begin{aligned}
    N_v^{(0)} &= 1\\
    m_v^{(0)} &\sim \mathcal{N}_d(0,1) \\ 
    e_v^{(0)} &= m_v^{(0)}, \\
\end{aligned}
\end{equation}
and updated with each codeword $e_v$ being the exponential moving average (EMA) of all the fine-grained representations $H = \left\{ f^M(h_{i,l}^M)~\middle|~\bar{h}_{i,l}^M = e_v \right\}$ that was replaced by $e_v$ for every training step $t$:
\begin{equation}
\begin{aligned}
    N_v^{(t)} &\leftarrow \gamma~N_v^{(t-1)} + (1-\gamma)~\left| H\right| \\
    m_v^{(t)} &\leftarrow \gamma~m_v^{(t-1)} + (1-\gamma)~\sum_{h\in H} h \\
    e_v^{(t)} &\leftarrow \frac{m_v^{(t)}}{N_v^{(t)}}, \\
\end{aligned}
\end{equation}
where the decay factor $\gamma$ is set to 0.99 throughout this work.
To improve the overall usage of the codebook, the input fine-grained representations are modality-wise batch normalized.
In addition, codewords that are not activated (i.e. $\left| H\right|=0$) for 100 consecutive steps are re-initialized during codebook update.
The reset value is randomly chosen from activated codewords.

\section{Implementation Details}
\label{sec:detail}

For each dataset and modality pair considered in this work, we selected baseline models that follow the cross-modal learning paradigm (as described in Section~\ref{subsec:paradigm}).
Baseline models with different fine-grained and high-level encoders ($f^M_\text{fine}$ and $f^M_\text{high}$) are summarized in Table~\ref{tb:setup}.
The links to the official implementation of these baseline models are also provided in the table.
For a fair comparison, we retrained the models with the $\mathcal{L}_\text{MMS}$ (margin set to 1e-3) as our baseline models.

\vspace{5pt}
\begin{table*}[h]
\centering
\begingroup
\renewcommand{\arraystretch}{1.4}
\begin{threeparttable}
\small
  \caption{Experiment setup on MSR-VTT, S-MiT, and Places.}
  \label{tb:setup}
  \centering
  \begin{tabular}{ l |  c  | l l  }
    \toprule
     Setup &  Modality & \multicolumn{2}{|c}{ Encoders from baseline model } \\
     \hline
     ~~Dataset & \textit{A}   &  \multicolumn{1}{c}{$f_\text{fine}^A$} & \multicolumn{1}{c}{$f_\text{high}^A$}  \\
     ~~- Baseline model & \textit{B}  &   \multicolumn{1}{c}{$f_\text{fine}^B$} & \multicolumn{1}{c}{$f_\text{high}^B$} \\
    \hline \hline
    ~~MSR-VTT~\cite{xu2016msr}& video  &   Vision Transformer\textsuperscript{4}~\cite{dosovitskiy2020image} & Avg. Pooling + Linear \\
    ~~- CLIP4Clip\textsuperscript{1}~\cite{luo2021clip4clip} & text  &  Transformer\textsuperscript{4}~\cite{vaswani2017attention,radford2019language} & \texttt{[EOT]} token + Linear  \\ \hline
    ~~S-MiT~\cite{monfort2021spoken} & video  & ResNet-152\textsuperscript{5}~\cite{He_2016_CVPR} + TSM\textsuperscript{6}~\cite{lin2019tsm} & Max Pooling + GLU~\cite{dauphin2017language}  \\
    ~~- AVLnet\textsuperscript{2}~\cite{rouditchenko2020avlnet} & audio   & Spectrogram+1D-ResNet~\cite{harwath2020jointly} & Avg. Pooling + GLU~\cite{dauphin2017language} \\ \hline
    ~~Places~\cite{harwath2017unsupervised} & image  & ResNet-50\textsuperscript{5}~\cite{He_2016_CVPR} & Avg. Pooling + GLU~\cite{dauphin2017language}\\
    ~~- ResDAVEnet\textsuperscript{3}~\cite{harwath2020jointly}   & audio  &  Spectrogram+1D-ResNet~\cite{harwath2020jointly} & Avg. Pooling + GLU~\cite{dauphin2017language} \\ \hline
    \bottomrule
  \end{tabular}
\begin{tablenotes}
\item{\textsuperscript{1}} \small{\url{https://github.com/ArrowLuo/CLIP4Clip}}
\item{\textsuperscript{2}}\small{\url{https://github.com/roudimit/AVLnet} (under BSD license)}
\item{\textsuperscript{3}} \small{\url{https://github.com/wnhsu/ResDAVEnet-VQ} (under BSD license)}
\item{\textsuperscript{4}} \small{\text{Initialized from CLIP model pretrained on WebImageText dataset~\cite{radford2021learning}.}}
\item{\textsuperscript{5}} \small{\text{Pretrained on ImageNet~\cite{deng2009imagenet}.}}
\item{\textsuperscript{6}} \small{\text{Pretrained on Multi-MiT~\cite{monfort2019multi}.}}

\end{tablenotes}
\end{threeparttable}
\vspace{5pt}
\endgroup
\end{table*}

\textbf{MSR-VTT.} For our baseline model, we did not reproduce CLIP4Clip's post-pretraining stage, which trained CLIP4Clip on the subset of HowTo100M~\cite{miech2019howto100m} before adapting to MSR-VTT, since this stage is not necessary for the best results on MSR-VTT and the subset is not released. 
We used all of the hyper-parameters of the official implementation except the batch size is reduced from 128 to 64 to meet our hardware restriction.
To train the shared discrete embedding space, we warm-started from the baseline model with a learning rate of 1e-5.
Each video is encoded into 8 codewords ($2\times 2\times 2$ for time, height, width) and each subword unit in the sentence is encoded into 1 codeword.
The baseline model is trained for 12 hours on 8 2080Ti GPUs; and it takes an additional 6 hours to train the proposed framework.

\textbf{S-MiT.} 
The input audio feature is a 40 dimensional mel-spectrogram with a window size of 25 ms and a hop size of 10 ms.
The baseline is trained with a batch size of 2048 and a learning rate of 1e-3.
To train the shared discrete embedding space, we warm-started from the baseline model with a learning rate of 1e-4.
Each video is encoded into 27 codewords ($3\times 3\times 3$ for time, height, width) and every 16 consecutive frames from the spectrogram is encoded into 1 codeword.
The baseline model is trained for 4 hours on 4 V100 GPUs; and it takes an additional 1 hour to train the proposed framework.
For both baseline model and our proposed model, we followed the previous work~\cite{monfort2021spoken} to perform a second round training with a learning rate of 1e-5 and a batch size of 128.
The second round training fine-tunes the TSM video encoder (which is frozen in the first round training) on S-MiT jointly with the rest of the components, which takes 2 days on 8 Titan RTX GPUs.

\textbf{Places.} 
The input audio feature is a 40 dimentional mel-spectrogram with a window size of 25 ms and a hop size of 10 ms.
The baseline is trained with a batch size of 256 and a learning rate of 1e-3.
To train the shared discrete embedding space, we warm-started from the baseline model with a learning rate of 1e-4.
Each image is encoded into 49 codewords ($7\times 7$ for height, width)  and every 16 consecutive frames from the spectrogram is encoded into 1 codeword.
The baseline model is trained for 36 hours on 1 V100 GPU; and it takes an additional 4 hours to train the proposed framework.

\section{MSR-VTT Video Retreival Full Comparison}
\label{sec:full_msrvtt}

\begin{table*}[h]
\centering
\begingroup
\setlength{\tabcolsep}{5pt}
\renewcommand{\arraystretch}{1.2}
\begin{threeparttable}
\small
  \caption{Full comparison against prior works on MSR-VTT text-to-video retrieval.}
  \label{tb:msrvtt}
  \centering
  \begin{tabular}{l | c c c c  }
    \toprule
    \multirow{3}{*}{~~~~~~Method} & \multicolumn{4}{c}{ Video Retrieval }  \\
      & \multicolumn{4}{c}{ (Text $\rightarrow$ Video) }   \\
     & R@1 $\uparrow$ & R@5 $\uparrow$ & R@10 $\uparrow$ & MnR $\downarrow$  \\
    \hline\hline
    ~~Collaborative Experts~\cite{liu2019use} & 20.9 & 48.8 & 62.4 & 28.2  \\
    ~~Multi-Modal Transformer~\cite{gabeur2020multi} & 26.6 & 57.1 & 69.6 & 24.0  \\
    ~~Support-Set Bottlenecks~\cite{patrick2020support} & 30.1 & 58.5 & 69.3 & -  \\
    ~~Multidomain Multimodal Transformer~\cite{dzabraev2021mdmmt} & 38.9 & 69.0 & 79.7 & 16.5  \\
    ~~Frozen-in-Time~\cite{bain2021frozen} & 31.0 & 59.5 & 70.5 & -  \\
    ~~Hierarchical Transformer with Momentum
Contrast~\cite{liu2021hit} & 30.7 & 60.9 & 73.2 & - \\
    ~~TeachText~\cite{croitoru2021teachtext} & 29.6 & 61.6 & 74.2 & -  \\
    ~~CLIP4Clip-meanPooling~\cite{luo2021clip4clip} & 43.1 & 70.4 & 80.8 & 16.2  \\
    ~~CLIP4Clip-seqLSTM~\cite{luo2021clip4clip} & 42.5 & 70.8 & 80.7 & 16.7  \\
    ~~CLIP4Clip-seqTransformer~\cite{luo2021clip4clip} & 44.5 & 71.4 & 81.6 & 15.3  \\
    ~~CLIP4Clip-tightTransformer~\cite{luo2021clip4clip} & 40.2 & 71.5 & 80.5 & 13.4  \\ \hline
    ~~Our Baseline (based on CLIP4Clip-meanPooling) & 42.6 & 71.2 & 80.8 & 15.5 \\
    ~~\textbf{Proposed} & 43.4 & 72.3 & 81.2 & 14.8  \\
    \bottomrule
  \end{tabular}
\end{threeparttable}
\endgroup
\end{table*}

In addition to the comparison against recent state-of-the-art methods in Table~\ref{tb:retrieve} for video retrieval on MSR-VTT, in Table~\ref{tb:msrvtt} we show the complete comparison to prior work and summarize the models here.
Collaborative Experts~\cite{liu2019use} leverages ``expert'' features that can be obtained from the raw video from different off-the-shelf models (such as object detection, scene classification, and speech recognition models) to build representations.
Instead of summarizing the expert features into a compact video representation and computing similarity with the text representation, the Multi-Modal Transformer~\cite{gabeur2020multi} computes similarity between different expert features and the text representation with a proposed variation of the Transformer~\cite{vaswani2017attention}.
Based on the Multi-Modal Transformer, Multidomain Multi-Modal Transformer~\cite{dzabraev2021mdmmt} explored an additional motion feature and the combination of different training datasets to further improve the result.
Support-Set Bottlenecks~\cite{patrick2020support} studies the benefit that cross-instance captioning can bring by generating text based on the combination of all representations of similar videos.
Similar to our framework, Hierarchical Transformer with Momentum Contrast~\cite{liu2021hit} divided representations from different layer of the encoders into fine-grained (which they referred to feature-level) and high-level (which they reffered to semantic-level) representations.
While our work focused on learning discrete representations in the fine-grained embedding space, they performed momentum-based representation matching across the two levels that encourages the two embedding spaces to be more similar.
TeachText~\cite{croitoru2021teachtext} leverages distillation learning where multiple captions describing the same video can be considered by different teacher models that jointly guide the student network.
Frozen-in-Time~\cite{bain2021frozen} and CLIP4Clip~\cite{luo2021clip4clip} both found the recent proposed Visual Transformer~\cite{dosovitskiy2020image} can significantly improve retrieval results while they differ in the choice of summarizing function for video (i.e. $f^M_\text{high}$) and the pre-training procedure.
Moreover, CLIP4Clip also introduces different choice of the summarizing function $f^M_\text{high}$ including RNNs (CLIP4Clip-seqLSTM) and Transformers (CLIP4Clip-seqTransformer) that replaces the mean-pooling function (CLIP4Clip-meanPooling) at the cost of higher time complexity and computational cost.
Note that while our work is based on the vanilla mean-pooling function, we achieved comparable or better performance with the proposed discrete embedding representations.
As described in Section~\ref{subsec:other_model}, CLIP4Clip also introduced a cross-modal transformer network (CLIP4Clip-tightTransformer) that allows cross-modal reference for deriving representations.

\section{Places Image Retrieval Full Comparison}
\label{sec:full_places}
\begin{table*}[h]
\centering
\begingroup
\setlength{\tabcolsep}{5pt}
\renewcommand{\arraystretch}{1.2}
\begin{threeparttable}
  \caption{\normalsize{Full comparison against prior works on Places image and spoken caption retrieval.}}
  \vspace{-0.3cm}
  \label{tb:places}
  \centering
  \small
  \begin{tabular}{l  c c c c  c c c c  }
    \toprule
     & \multicolumn{4}{c}{ Audio to Image}  & \multicolumn{4}{c}{ Image to Audio} \\
     & R@1 $\uparrow$ & R@5 $\uparrow$ & R@10 $\uparrow$ & MnR $\downarrow$ & R@1 $\uparrow$ & R@5 $\uparrow$ & R@10 $\uparrow$ & MnR $\downarrow$  \\
    \hline
    ~~Harwath~et~al.~\cite{harwath2017unsupervised}\textdaggerdbl & 14.8 & 40.3 & 54.8 & -- & 12.1 & 33.5 & 46.3 & -- \\
    ~~Harwath~et~al.~\cite{harwath2017learning}\textdaggerdbl & 16.1 & 40.4 & 56.4 & -- & 13.0 & 37.8 & 54.2& -- \\
    ~~DAVEnet~\cite{harwath2018jointly} & 20.0 & 46.9 & 60.4 & -- & 12.7 & 37.5 & 52.8 & -- \\
    ~~ResDAVEnet~\cite{harwath2020jointly}* & 30.9 & 63.6 & 74.2 & 20.2 & 26.4 & 58.5 & 71.2 & 21.6 \\
    ~~ResDAVEnet-VQ~\cite{Harwath2020Learning}* & 34.9 & 70.2 & 79.4 & 15.0 & 32.7 & 65.6 & 77.0 & 18.0 \\
    ~~Our Baseline{\textdagger} & 43.8 & 74.1 & 82.4 & 15.8 & 40.4 & 73.3 & 82.5 & 10.9 \\
    ~~\textbf{Proposed} & \textbf{46.5} & \textbf{77.4} & \textbf{85.8} & \textbf{13.7} & \textbf{45.4} & \textbf{77.7} & \textbf{85.9}  &  \textbf{8.9}\\
    \bottomrule
  \end{tabular}
\begin{tablenotes}
\item{\textdaggerdbl} \small{\text{Results found in~\cite{harwath2018jointly}.}}
\item{\textdagger} \small{\text{Existing model reproduced with $\mathcal{L}_\text{MMS}$ for fair comparison.}}
\item{*} \small{\text{Results obtained by running the official code and pre-trained models.}}
\end{tablenotes}
\end{threeparttable}
\vspace{-10pt}
\endgroup
\end{table*}
We show the full comparison to prior work on Places-400k in Table~\ref{tb:places}.
The previous methods~\cite{harwath2017learning,harwath2017unsupervised,harwath2018jointly} use less complex audio and image encoders with fewer parameters.

\section{Results Without Cross-Modal Code Matching}
\label{sec:no_mcmc}
\begin{figure*}[h]
\begin{center}
\centerline{\includegraphics[width=\linewidth]{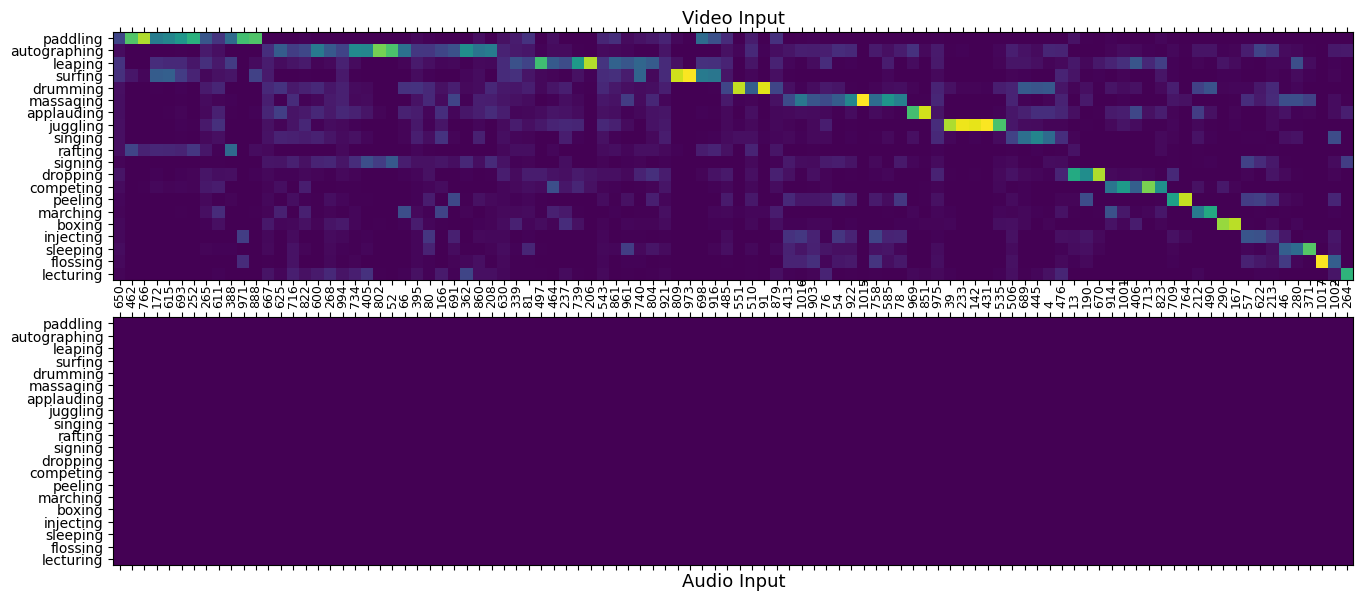}}
\end{center}
\vspace{-20pt}
\caption{Conditional probability matrix between codewords and action labels learned by our proposed method when the Cross-Modal Code Matching objective is excluded.}
\label{fig:cor_zero}
\end{figure*}
\begin{figure*}[h]
\begin{center}
\centerline{\includegraphics[width=\linewidth]{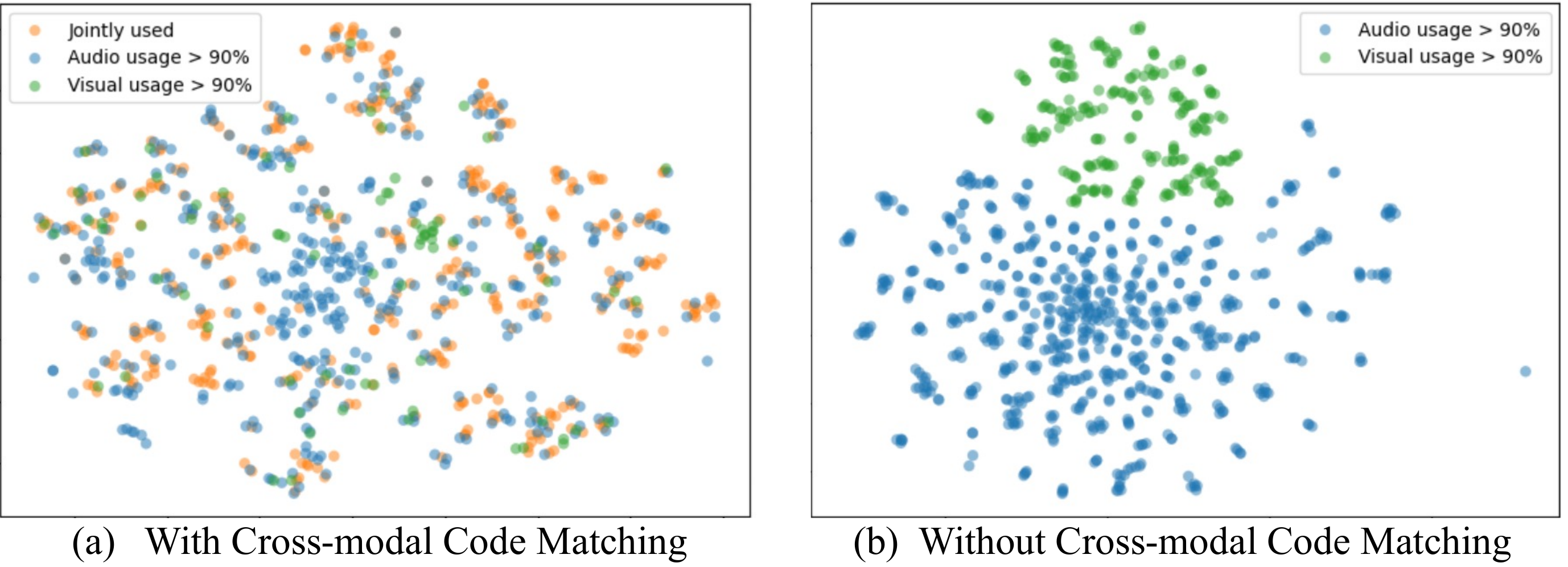}}
\end{center}
\vspace{-20pt}
\caption{T-SNE visualization of the codebook with and without the proposed \mmloss~Objective. Each point corresponds to a codeword colored with respect to the input modality that utilized it the most. Codewords without high ($>$ 90\%) usage from single modality are labeled as ``jointly used''. }
\label{fig:tsne}
\end{figure*}

To demonstrate the importance of our proposed Cross-Modal Code Matching objective, Figure~\ref{fig:cor_zero} illustrates the conditional probability matrix (described in Section~\ref{subsec:codeword_exp} and Figure~\ref{fig:cor}) when the proposed objective is deactived (setting $\alpha = 0$).
Unsurprisingly, we see that the correlation between codewords and action labels are gone, indicating that the assignment of codewords are now dominated by the input modality instead of the underlying action label.
This can also be verified by visualizing the discrete embedding space in a lower dimension as plotted in Figure~\ref{fig:tsne}.
This evidence suggests that the proposed \mmloss~Objective is effective for learning modality-invariant representations.

\section{Additional Codeword Correspondence and Localization Examples}
\label{sec:more_demo}

\begin{figure*}[h]
\begin{center}
\centerline{\includegraphics[width=0.6
\linewidth]{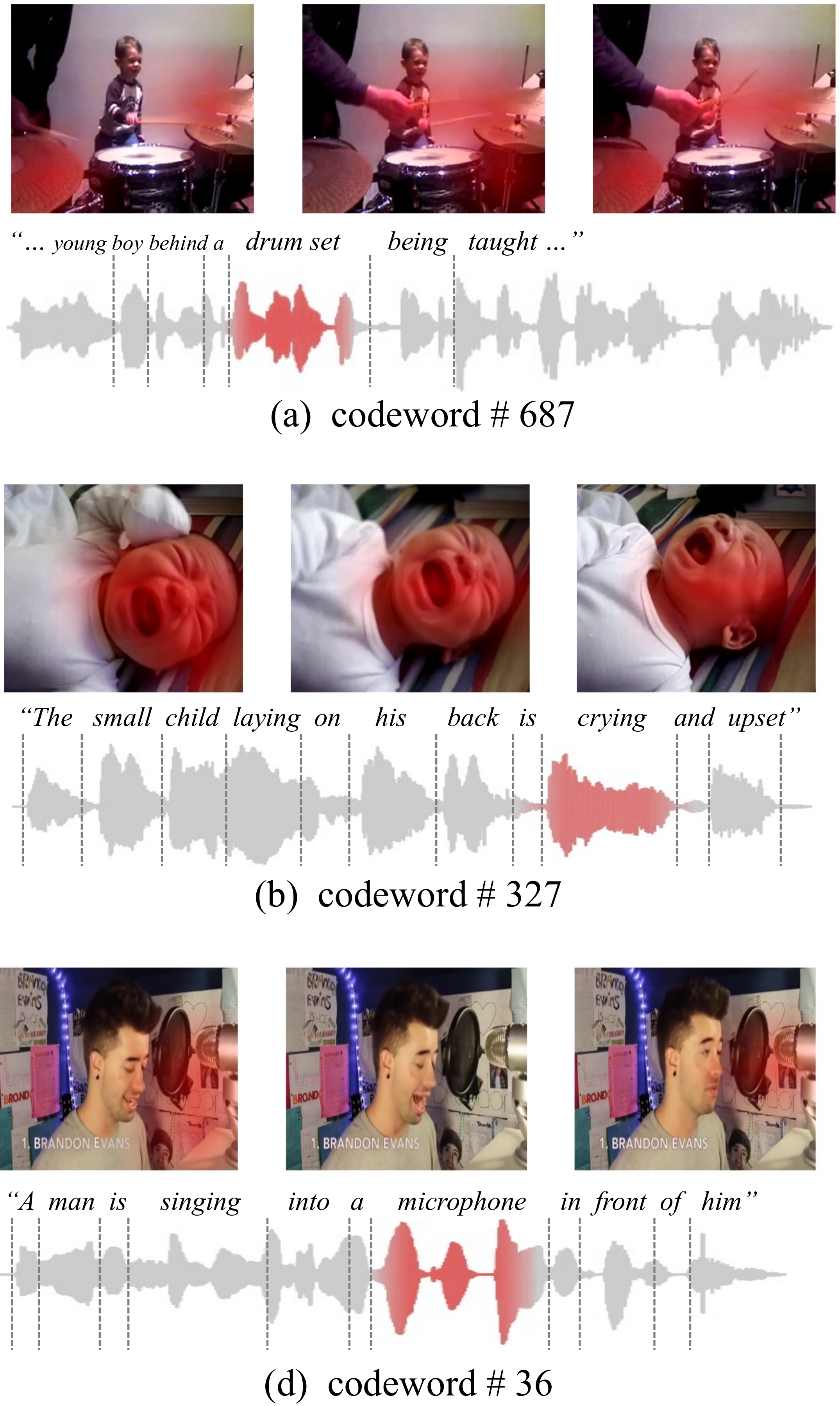}}
\end{center}
\caption{More examples for codeword cross-modal localization.}
\label{fig:demo_full}
\end{figure*}

An extension of Table~\ref{tb:cooccur} showing the correspondence between codewords, visual actions, and spoken words are provided in Table~\ref{tb:cooccur_full}.
We also provide more examples for codeword localization in Figure~\ref{fig:demo_full}.

\newpage
\begin{table*}
\centering
\begingroup
\setlength{\tabcolsep}{5pt}
\renewcommand{\arraystretch}{1.2}
\scriptsize
  \caption{Correspondence between codewords, visual actions, and spoken words (Extended  Table~\ref{tb:cooccur}). The second hypothesis and the occurrence are omitted for simplicity. All codewords activated on S-MiT's development set are listed.}
  \label{tb:cooccur_full}
  \centering
  \begin{tabular}{c  c  c c  c c  C  c  c c  c c }
    \toprule
     \multirow{3}{*}{Rank} &\multirow{3}{*}{Code}  &  \multicolumn{2}{c}{ Visual Action }  & \multicolumn{2}{c}{ Spoken word } &  \multirow{3}{*}{Rank} &\multirow{3}{*}{Code}  &  \multicolumn{2}{c}{ Visual Action }  & \multicolumn{2}{c}{ Spoken word }\\
      & &\multicolumn{2}{c}{ Top Hypothesis }  &  \multicolumn{2}{c}{ Top Hypothesis } & & &\multicolumn{2}{c}{ Top Hypothesis }  &  \multicolumn{2}{c}{ Top Hypothesis }\\
     & &label & Prc.  & word & F1 && &label & Prc.  & word & F1  \\
    \cline{1-6}\cline{7-12}
1 & 201 & juggling & 97.5 & juggling & 36.7 & 61 & 940 & landing & 44.9 & airplane & 19.6\\
2 & 349 & flossing & 96.0 & floss & 15.8 & 62 & 262 & sewing & 44.7 & sewing & 13.2\\
3 & 145 & surfing & 95.6 & surfboard & 23.7 & 63 & 532 & autographing & 44.4 & selfie & 22.2\\
4 & 29 & tattooing & 94.6 & tattoo & 15.8 & 64 & 928 & stirring & 44.1 & boiling & 27.3\\
5 & 233 & ironing & 93.8 & ironing & 20.5 & 65 & 747 & applauding & 43.8 & clapping & 23.8\\
6 & 766 & surfing & 93.2 & surfing & 22.1 & 66 & 447 & paddling & 43.1 & boat & 8.3\\
7 & 191 & juggling & 90.2 & juggling & 29.1 & 67 & 823 & skipping & 43.0 & jump & 17.1\\
8 & 753 & autographing & 85.0 & autographs & 26.4 & 68 & 308 & shaving & 42.5 & comb & 10.0\\
9 & 606 & autographing & 83.7 & signing & 16.2 & 69 & 518 & skiing & 41.8 & skiing & 11.4\\
10 & 640 & drumming & 81.6 & drums & 19.5 & 70 & 860 & bulldozing & 41.7 & bulldozer & 25.7\\
11 & 436 & injecting & 81.6 & injected & 13.2 & 71 & 61 & extinguishing & 41.3 & sting & 9.1\\
12 & 109 & peeling & 80.9 & peeling & 21.2 & 72 & 296 & combing & 40.9 & brushes & 5.9\\
13 & 551 & shaving & 80.2 & shaving & 18.0 & 73 & 435 & screwing & 40.8 & drill & 25.0\\
14 & 137 & paddling & 80.0 & canoe & 25.8 & 74 & 705 & surfing & 40.6 & ocean & 27.0\\
15 & 327 & crying & 78.8 & crying & 29.5 & 75 & 760 & hammering & 40.0 & hammering & 23.3\\
16 & 593 & surfing & 77.7 & surfboard & 10.9 & 76 & 926 & paddling & 40.0 & lake & 6.8\\
17 & 687 & drumming & 77.3 & drums & 14.4 & 77 & 888 & paddling & 39.6 & lake & 7.4\\
18 & 883 & tattooing & 77.2 & tattoo & 13.6 & 78 & 169 & dunking & 39.3 & nba & 7.5\\
19 & 1000 & inflating & 74.5 & inflatable & 12.8 & 79 & 681 & manicuring & 38.7 & nails & 13.2\\
20 & 222 & boxing & 71.3 & boxing & 13.2 & 80 & 685 & signing & 38.6 & writing & 8.3\\
21 & 243 & shredding & 70.0 & shredding & 28.6 & 81 & 631 & paddling & 38.5 & clouds & 12.2\\
22 & 157 & paddling & 69.9 & kayak & 21.3 & 82 & 800 & dropping & 38.3 & beans & 12.9\\
23 & 427 & boxing & 69.8 & boxers & 16.2 & 83 & 556 & drumming & 38.3 & marching & 11.6\\
24 & 774 & surfing & 69.2 & waves & 23.0 & 84 & 758 & wrapping & 38.1 & wrapping & 22.2\\
25 & 613 & manicuring & 67.9 & nails & 24.5 & 85 & 368 & texting & 38.0 & texting & 16.7\\
26 & 952 & leaping & 66.0 & dolphins & 10.7 & 86 & 625 & combing & 37.9 & hair & 4.9\\
27 & 196 & boxing & 64.1 & boxer & 13.9 & 87 & 166 & boxing & 37.8 & boxing & 7.2\\
28 & 706 & sailing & 63.4 & sailboat & 18.8 & 88 & 539 & paddling & 37.5 & helmet & 13.0\\
29 & 58 & shaving & 62.8 & shaving & 10.9 & 89 & 139 & leaping & 37.5 & jumping & 16.6\\
30 & 759 & paddling & 60.7 & paddling & 12.4 & 90 & 123 & drumming & 37.1 & playing & 8.7\\
31 & 868 & boxing & 60.0 & boxer & 11.2 & 91 & 577 & drumming & 37.0 & musical & 8.1\\
32 & 500 & dialing & 60.0 & dialing & 13.8 & 92 & 780 & screwing & 36.9 & drill & 15.8\\
33 & 536 & cheering & 60.0 & cheerleaders & 26.8 & 93 & 621 & leaping & 36.6 & jumps & 9.7\\
34 & 50 & rafting & 58.6 & rafting & 16.7 & 94 & 154 & boxing & 36.0 & referee & 14.7\\
35 & 664 & dunking & 58.0 & basketball & 11.0 & 95 & 415 & grilling & 35.7 & grill & 15.7\\
36 & 103 & autographing & 57.8 & carpet & 8.2 & 96 & 345 & autographing & 35.5 & pictures & 19.3\\
37 & 990 & wrestling & 56.1 & wrestling & 25.9 & 97 & 694 & sailing & 34.9 & sailing & 7.0\\
38 & 880 & sleeping & 56.0 & sleeping & 21.1 & 98 & 973 & leaping & 34.4 & tale & 8.0\\
39 & 48 & paddling & 55.1 & rowing & 18.2 & 99 & 957 & shrugging & 34.4 & lifting & 10.3\\
40 & 292 & skiing & 54.2 & skiing & 20.0 & 100 & 713 & paddling & 34.3 & sunset & 25.3\\
41 & 602 & ironing & 52.5 & ironing & 7.1 & 101 & 697 & injecting & 34.1 & doctor & 18.8\\
42 & 954 & dropping & 52.4 & dropped & 8.2 & 102 & 431 & peeling & 33.9 & apple & 20.0\\
43 & 735 & applauding & 52.1 & clapping & 23.4 & 103 & 164 & typing & 33.8 & laptop & 20.6\\
44 & 816 & autographing & 51.0 & carpet & 22.5 & 104 & 776 & juggling & 33.6 & balls & 16.5\\
45 & 516 & swinging & 50.0 & swing & 20.4 & 105 & 73 & shrugging & 32.9 & weight & 14.6\\
46 & 421 & carving & 50.0 & carving & 27.2 & 106 & 846 & injecting & 32.8 & gloves & 7.8\\
47 & 168 & drumming & 49.3 & marching & 17.5 & 107 & 395 & juggling & 32.7 & balls & 10.1\\
48 & 561 & flossing & 48.0 & mouse & 10.0 & 108 & 273 & dusting & 32.6 & clean & 11.5\\
49 & 970 & marrying & 47.8 & bride & 22.2 & 109 & 737 & paddling & 32.5 & mountains & 14.0\\
50 & 610 & dunking & 47.4 & basketball & 19.5 & 110 & 291 & coughing & 32.4 & sneezes & 15.6\\
51 & 105 & paddling & 47.2 & river & 23.7 & 111 & 375 & colliding & 32.4 & crashing & 14.5\\
52 & 150 & waxing & 47.2 & wax & 20.3 & 112 & 693 & sleeping & 32.3 & baby & 28.9\\
53 & 92 & howling & 46.7 & barking & 15.1 & 113 & 111 & baking & 32.3 & baker & 13.8\\
54 & 929 & typing & 46.3 & typing & 22.4 & 114 & 805 & massaging & 32.0 & squatted & 8.7\\
55 & 844 & drumming & 46.2 & band & 14.5 & 115 & 134 & autographing & 31.7 & obama & 7.5\\
56 & 497 & cheering & 45.8 & cheerleaders & 34.8 & 116 & 923 & wrapping & 31.6 & tape & 16.7\\
57 & 322 & paddling & 45.8 & kayak & 7.2 & 117 & 698 & surfing & 31.5 & beach & 9.8\\
58 & 672 & boxing & 45.6 & fighting & 28.8 & 118 & 362 & paddling & 31.5 & water & 8.2\\
59 & 97 & barbecuing & 45.6 & grill & 26.4 & 119 & 505 & drumming & 31.0 & guitar & 13.1\\
60 & 216 & inflating & 45.3 & balloon & 10.3 & 120 & 215 & shaving & 31.0 & vent & 12.1\\

    \bottomrule
  \end{tabular}
\endgroup
\end{table*}
\setcounter{table}{5}
\begin{table*}
\centering
\begingroup
\setlength{\tabcolsep}{5pt}
\renewcommand{\arraystretch}{1.2}
\scriptsize
  \caption{continued}
  \centering
  \begin{tabular}{c  c  c c  c c  C  c  c c  c c }
    \toprule
         \multirow{3}{*}{Rank} &\multirow{3}{*}{Code}  &  \multicolumn{2}{c}{ Visual Action }  & \multicolumn{2}{c}{ Spoken word } &  \multirow{3}{*}{Rank} &\multirow{3}{*}{Code}  &  \multicolumn{2}{c}{ Visual Action }  & \multicolumn{2}{c}{ Spoken word }\\
      & &\multicolumn{2}{c}{ Top Hypothesis }  &  \multicolumn{2}{c}{ Top Hypothesis } & & &\multicolumn{2}{c}{ Top Hypothesis }  &  \multicolumn{2}{c}{ Top Hypothesis }\\
     & &label & Prc.  & word & F1 && &label & Prc.  & word & F1  \\
    \cline{1-6}\cline{7-12}
121 & 642 & autographing & 30.8 & sign & 9.9 & 181 & 646 & autographing & 23.5 & taking & 11.6\\
122 & 828 & paddling & 30.6 & river & 3.6 & 182 & 423 & applauding & 23.5 & crowd & 11.1\\
123 & 6 & leaping & 30.3 & monkey & 31.4 & 183 & 699 & racing & 23.4 & motorcycle & 16.5\\
124 & 974 & sprinkling & 30.0 & sprinkler & 26.7 & 184 & 651 & paddling & 23.4 & sky & 4.5\\
125 & 44 & flossing & 29.9 & teeth & 3.0 & 185 & 414 & drenching & 23.3 & rain & 16.9\\
126 & 342 & drumming & 29.9 & playing & 7.7 & 186 & 55 & racing & 23.3 & race & 12.0\\
127 & 108 & boxing & 29.8 & practicing & 24.1 & 187 & 718 & drumming & 23.2 & costume & 9.2\\
128 & 784 & pedaling & 29.7 & bikes & 13.1 & 188 & 439 & pedaling & 23.1 & cyclist & 12.6\\
129 & 266 & barbecuing & 29.7 & meat & 22.5 & 189 & 19 & clipping & 23.1 & tractor & 22.2\\
130 & 991 & drumming & 29.6 & guitar & 11.0 & 190 & 255 & paddling & 23.1 & water & 3.6\\
131 & 597 & signing & 29.3 & writing & 4.7 & 191 & 701 & lecturing & 23.0 & preacher & 16.3\\
132 & 817 & welding & 29.1 & steel & 11.6 & 192 & 444 & autographing & 22.9 & protesters & 7.4\\
133 & 673 & typing & 29.1 & laptop & 12.3 & 193 & 859 & singing & 22.9 & performer & 5.6\\
134 & 113 & dialing & 29.0 & telephone & 11.7 & 194 & 18 & applauding & 22.9 & cheering & 16.2\\
135 & 470 & sawing & 28.9 & saw & 10.5 & 195 & 371 & barbecuing & 22.9 & fire & 10.1\\
136 & 657 & landing & 28.7 & airplane & 11.7 & 196 & 315 & peeling & 22.8 & orange & 19.9\\
137 & 440 & surfing & 28.6 & cap & 6.1 & 197 & 271 & racing & 22.7 & race & 11.1\\
138 & 404 & rinsing & 28.6 & scrubbing & 13.3 & 198 & 955 & leaping & 22.6 & seagulls & 24.2\\
139 & 0 & applauding & 28.6 & protesting & 13.8 & 199 & 584 & boxing & 22.6 & bag & 23.7\\
140 & 950 & paddling & 28.2 & water & 9.7 & 200 & 555 & pitching & 22.5 & baseball & 19.6\\
141 & 430 & hiking & 27.8 & hikers & 13.8 & 201 & 286 & piloting & 22.5 & helicopter & 12.5\\
142 & 762 & leaping & 27.8 & diving & 12.0 & 202 & 569 & paddling & 22.3 & down & 17.6\\
143 & 504 & bowing & 27.3 & praying & 19.0 & 203 & 692 & paddling & 22.2 & train & 31.4\\
144 & 295 & paddling & 27.2 & bridge & 26.4 & 204 & 682 & paddling & 22.1 & trees & 16.3\\
145 & 579 & dunking & 27.2 & ball & 10.5 & 205 & 116 & slicing & 22.0 & cutting & 22.4\\
146 & 380 & leaping & 26.7 & deer & 29.3 & 206 & 442 & dropping & 22.0 & wipers & 16.3\\
147 & 152 & sleeping & 26.7 & laying & 14.3 & 207 & 324 & skiing & 22.0 & skis & 4.1\\
148 & 603 & leaping & 26.5 & slipping & 4.7 & 208 & 924 & flooding & 21.9 & flooded & 16.3\\
149 & 838 & dusting & 26.5 & vacuum & 14.3 & 209 & 826 & bulldozing & 21.6 & tractor & 7.0\\
150 & 825 & scooping & 25.9 & spilled & 16.7 & 210 & 422 & falling & 21.4 & waterfall & 19.4\\
151 & 64 & pedaling & 25.9 & bicycles & 8.5 & 211 & 931 & bulldozing & 21.4 & bulldozer & 18.2\\
152 & 455 & erupting & 25.6 & smoke & 20.6 & 212 & 259 & wrestling & 21.3 & cuddling & 8.0\\
153 & 429 & competing & 25.5 & field & 13.0 & 213 & 475 & leaping & 21.2 & dance & 6.1\\
154 & 989 & competing & 25.5 & football & 19.0 & 214 & 905 & jumping & 21.2 & horse & 29.1\\
155 & 223 & competing & 25.4 & soccer & 25.0 & 215 & 806 & jogging & 21.2 & jogging & 14.3\\
156 & 51 & bowling & 25.4 & dome & 8.2 & 216 & 813 & applauding & 21.1 & waving & 15.9\\
157 & 379 & slicing & 25.4 & slicing & 12.2 & 217 & 538 & paddling & 21.0 & water & 6.7\\
158 & 911 & paddling & 25.4 & aerial & 28.0 & 218 & 101 & massaging & 20.9 & dog & 13.5\\
159 & 364 & leaping & 25.4 & bed & 18.6 & 219 & 482 & swinging & 20.9 & swinging & 7.9\\
160 & 483 & paddling & 25.3 & flowing & 5.7 & 220 & 680 & leaping & 20.9 & air & 24.1\\
161 & 634 & autographing & 25.0 & graduation & 4.4 & 221 & 1018 & dialing & 20.7 & tapping & 44.4\\
162 & 884 & leaping & 25.0 & trampoline & 8.8 & 222 & 665 & shaving & 20.7 & hair & 4.1\\
163 & 485 & stirring & 25.0 & pan & 20.3 & 223 & 417 & drumming & 20.6 & stage & 8.1\\
164 & 540 & boxing & 25.0 & jacks & 6.7 & 224 & 165 & mowing & 20.6 & lawn & 16.5\\
165 & 13 & paddling & 25.0 & boat & 18.1 & 225 & 194 & flossing & 20.6 & scoop & 6.9\\
166 & 873 & paddling & 25.0 & mountains & 8.9 & 226 & 200 & smashing & 20.5 & smashed & 12.2\\
167 & 909 & autographing & 24.3 & book & 14.0 & 227 & 453 & carving & 20.4 & wood & 17.5\\
168 & 638 & autographing & 24.3 & either & 3.3 & 228 & 57 & child+singing & 20.2 & singing & 18.5\\
169 & 963 & plugging & 24.3 & plug & 11.8 & 229 & 420 & paddling & 20.0 & forest & 13.3\\
170 & 131 & paddling & 24.2 & yellow & 26.5 & 230 & 918 & massaging & 19.8 & laying & 13.5\\
171 & 799 & welding & 24.2 & construction & 27.9 & 231 & 810 & paddling & 19.8 & dolphin & 2.9\\
172 & 486 & hammering & 24.1 & hammering & 6.0 & 232 & 520 & sailing & 19.7 & boats & 5.8\\
173 & 465 & competing & 24.0 & teams & 11.9 & 233 & 190 & knitting & 19.6 & string & 10.9\\
174 & 67 & lecturing & 24.0 & conference & 9.8 & 234 & 1016 & mopping & 19.6 & mopping & 15.1\\
175 & 325 & texting & 24.0 & phone & 12.7 & 235 & 317 & dunking & 19.4 & basket & 18.9\\
176 & 1001 & competing & 23.9 & soccer & 8.1 & 236 & 827 & paddling & 19.3 & ski & 8.7\\
177 & 242 & competing & 23.9 & football & 6.7 & 237 & 24 & leaping & 19.2 & dancing & 10.2\\
178 & 714 & calling & 23.7 & telephone & 6.7 & 238 & 1019 & dropping & 19.1 & falls & 12.1\\
179 & 89 & competing & 23.6 & soccer & 17.7 & 239 & 997 & sleeping & 19.0 & baby & 8.3\\
180 & 1013 & paddling & 23.5 & forest & 19.1 & 240 & 77 & peeling & 19.0 & makeup & 17.0\\

    \bottomrule
  \end{tabular}
\endgroup
\end{table*}
\setcounter{table}{5}
\begin{table*}
\centering
\begingroup
\setlength{\tabcolsep}{5pt}
\renewcommand{\arraystretch}{1.2}
\scriptsize
  \caption{continued}
  \centering
  \begin{tabular}{c  c  c c  c c  C  c  c c  c c }
    \toprule
         \multirow{3}{*}{Rank} &\multirow{3}{*}{Code}  &  \multicolumn{2}{c}{ Visual Action }  & \multicolumn{2}{c}{ Spoken word } &  \multirow{3}{*}{Rank} &\multirow{3}{*}{Code}  &  \multicolumn{2}{c}{ Visual Action }  & \multicolumn{2}{c}{ Spoken word }\\
      & &\multicolumn{2}{c}{ Top Hypothesis }  &  \multicolumn{2}{c}{ Top Hypothesis } & & &\multicolumn{2}{c}{ Top Hypothesis }  &  \multicolumn{2}{c}{ Top Hypothesis }\\
     & &label & Prc.  & word & F1 && &label & Prc.  & word & F1  \\
    \cline{1-6}\cline{7-12}
241 & 126 & leaping & 18.9 & exercising & 18.8 & 301 & 459 & paddling & 16.3 & view & 9.1\\
242 & 449 & leaping & 18.9 & tree & 17.6 & 302 & 323 & shaving & 16.3 & head & 19.8\\
243 & 187 & surfing & 18.9 & riding & 12.3 & 303 & 522 & dunking & 16.3 & court & 12.5\\
244 & 117 & raining & 18.8 & traffic & 21.7 & 304 & 773 & storming & 16.2 & storm & 9.8\\
245 & 671 & paddling & 18.8 & city & 13.8 & 305 & 748 & autographing & 16.2 & sidewalk & 14.3\\
246 & 736 & autographing & 18.7 & howling & 4.9 & 306 & 299 & punting & 16.2 & kicks & 7.3\\
247 & 251 & surfing & 18.5 & scuba & 7.0 & 307 & 981 & paddling & 16.2 & jacket & 14.1\\
248 & 491 & raining & 18.4 & simpsons & 8.5 & 308 & 627 & singing & 16.2 & dark & 14.4\\
249 & 1 & burying & 18.4 & dirt & 19.3 & 309 & 239 & fishing & 16.2 & fishing & 21.5\\
250 & 188 & autographing & 18.4 & beard & 8.8 & 310 & 41 & leaping & 16.1 & slow & 26.0\\
251 & 742 & pedaling & 18.3 & bike & 22.0 & 311 & 479 & leaping & 16.1 & kids & 6.8\\
252 & 531 & chewing & 18.2 & eats & 12.5 & 312 & 348 & reaching & 16.0 & slipping & 7.7\\
253 & 130 & applauding & 18.1 & crowd & 7.4 & 313 & 63 & dropping & 16.0 & leaves & 18.2\\
254 & 246 & clinging & 18.0 & bird & 31.2 & 314 & 892 & applauding & 16.0 & flag & 13.6\\
255 & 318 & dialing & 17.9 & phone & 6.8 & 315 & 558 & stirring & 16.0 & cooking & 9.9\\
256 & 329 & extinguishing & 17.9 & fire & 14.5 & 316 & 691 & paddling & 16.0 & background & 19.0\\
257 & 387 & barbecuing & 17.9 & sausages & 10.7 & 317 & 319 & leaping & 15.9 & up & 3.8\\
258 & 993 & autographing & 17.9 & movie & 7.6 & 318 & 845 & stirring & 15.8 & blade & 6.7\\
259 & 961 & paddling & 17.9 & rushing & 8.3 & 319 & 801 & paddling & 15.8 & mask & 14.7\\
260 & 921 & surfing & 17.8 & beach & 15.0 & 320 & 726 & swimming & 15.8 & swimming & 12.2\\
261 & 208 & cheering & 17.8 & stadium & 15.0 & 321 & 458 & shrugging & 15.8 & karate & 3.5\\
262 & 650 & leaping & 17.8 & jumps & 6.4 & 322 & 912 & applauding & 15.7 & old & 11.0\\
263 & 388 & dropping & 17.8 & float & 5.6 & 323 & 648 & peeling & 15.7 & kitchen & 13.8\\
264 & 78 & paddling & 17.8 & walnut & 6.5 & 324 & 572 & dialing & 15.5 & block & 3.2\\
265 & 332 & dropping & 17.7 & falling & 8.8 & 325 & 330 & paddling & 15.5 & waterfall & 3.3\\
266 & 244 & lecturing & 17.6 & giving & 7.3 & 326 & 211 & leaping & 15.5 & cat & 17.9\\
267 & 948 & paddling & 17.6 & across & 8.9 & 327 & 752 & paddling & 15.5 & trail & 6.7\\
268 & 1008 & surfing & 17.6 & scuba & 4.1 & 328 & 34 & sleeping & 15.5 & bed & 8.6\\
269 & 554 & sewing & 17.6 & machine & 12.2 & 329 & 792 & autographing & 15.5 & sitting & 3.5\\
270 & 604 & leaping & 17.6 & fish & 25.2 & 330 & 588 & sowing & 15.4 & farmer & 10.5\\
271 & 587 & saluting & 17.5 & soldier & 12.0 & 331 & 869 & pouring & 15.4 & poured & 20.5\\
272 & 509 & discussing & 17.5 & office & 23.5 & 332 & 840 & leaping & 15.4 & pool & 11.3\\
273 & 720 & competing & 17.5 & track & 24.2 & 333 & 407 & measuring & 15.4 & drawing & 8.5\\
274 & 1022 & shrugging & 17.5 & gym & 18.1 & 334 & 667 & welding & 15.4 & metal & 17.6\\
275 & 987 & autographing & 17.4 & baseball & 21.3 & 335 & 661 & colliding & 15.4 & hockey & 25.0\\
276 & 294 & drumming & 17.4 & stick & 16.9 & 336 & 560 & flossing & 15.4 & animation & 14.0\\
277 & 552 & applauding & 17.4 & crowd & 18.4 & 337 & 149 & lecturing & 15.4 & graphs & 8.7\\
278 & 995 & draining & 17.4 & waterfall & 14.3 & 338 & 175 & autographing & 15.3 & walking & 7.1\\
279 & 284 & drumming & 17.3 & concert & 29.3 & 339 & 815 & sleeping & 15.3 & baby & 17.5\\
280 & 808 & draining & 17.3 & water & 5.8 & 340 & 608 & autographing & 15.3 & people & 6.1\\
281 & 977 & snowing & 17.2 & snowy & 10.3 & 341 & 795 & leaping & 15.2 & animals & 13.6\\
282 & 495 & unloading & 17.1 & time-lapse & 21.7 & 342 & 755 & peeling & 15.2 & kitchen & 12.0\\
283 & 184 & autographing & 17.1 & hat & 16.8 & 343 & 138 & juggling & 15.2 & shirtless & 18.2\\
284 & 210 & paddling & 17.1 & rocks & 14.1 & 344 & 496 & hanging & 15.2 & hanging & 17.9\\
285 & 120 & boxing & 17.0 & shorts & 13.7 & 345 & 641 & competing & 15.1 & marching & 4.4\\
286 & 914 & paddling & 17.0 & two & 8.3 & 346 & 2 & drumming & 15.0 & stage & 16.7\\
287 & 263 & dropping & 16.9 & fruits & 8.1 & 347 & 916 & paddling & 15.0 & sunny & 3.4\\
288 & 245 & competing & 16.9 & kicking & 7.6 & 348 & 393 & chewing & 15.0 & eating & 15.4\\
289 & 639 & autographing & 16.8 & dress & 8.8 & 349 & 609 & autographing & 15.0 & talking & 2.5\\
290 & 739 & autographing & 16.7 & greenfield & 9.8 & 350 & 269 & draining & 15.0 & water & 12.2\\
291 & 704 & leaping & 16.7 & dance & 15.0 & 351 & 731 & flossing & 15.0 & demonstrating & 9.0\\
292 & 562 & splashing & 16.7 & splashes & 19.0 & 352 & 513 & dialing & 14.9 & finger & 10.0\\
293 & 658 & splashing & 16.7 & bottle & 24.6 & 353 & 382 & paddling & 14.9 & sky & 7.0\\
294 & 629 & sleeping & 16.7 & reports & 14.8 & 354 & 645 & autographing & 14.8 & people & 3.3\\
295 & 1014 & massaging & 16.6 & getting & 10.3 & 355 & 553 & juggling & 14.8 & spinning & 13.8\\
296 & 503 & pedaling & 16.5 & jogging & 7.2 & 356 & 490 & spitting & 14.8 & drink & 19.5\\
297 & 686 & paddling & 16.4 & nuts & 3.6 & 357 & 807 & crushing & 14.7 & crushed & 13.6\\
298 & 734 & singing & 16.4 & singing & 15.9 & 358 & 412 & autographing & 14.7 & player & 9.1\\
299 & 334 & autographing & 16.3 & papers & 6.8 & 359 & 900 & leaping & 14.7 & branch & 23.8\\
300 & 788 & signing & 16.3 & reading & 8.7 & 360 & 622 & paddling & 14.5 & rocks & 15.9\\
    \bottomrule
  \end{tabular}
\endgroup
\end{table*}
\setcounter{table}{5}
\begin{table*}
\centering
\begingroup
\setlength{\tabcolsep}{5pt}
\renewcommand{\arraystretch}{1.2}
\scriptsize
  \caption{continued}
  \centering
  \begin{tabular}{c  c  c c  c c  C  c  c c  c c }
    \toprule
         \multirow{3}{*}{Rank} &\multirow{3}{*}{Code}  &  \multicolumn{2}{c}{ Visual Action }  & \multicolumn{2}{c}{ Spoken word } &  \multirow{3}{*}{Rank} &\multirow{3}{*}{Code}  &  \multicolumn{2}{c}{ Visual Action }  & \multicolumn{2}{c}{ Spoken word }\\
      & &\multicolumn{2}{c}{ Top Hypothesis }  &  \multicolumn{2}{c}{ Top Hypothesis } & & &\multicolumn{2}{c}{ Top Hypothesis }  &  \multicolumn{2}{c}{ Top Hypothesis }\\
     & &label & Prc.  & word & F1 && &label & Prc.  & word & F1  \\
    \cline{1-6}\cline{7-12}
361 & 623 & paddling & 14.5 & yellow & 8.7 & 421 & 282 & dropping & 12.7 & backdrop & 8.9\\
362 & 462 & leaping & 14.5 & dancing & 18.5 & 422 & 443 & frying & 12.7 & food & 15.1\\
363 & 65 & autographing & 14.5 & pen & 5.0 & 423 & 676 & rinsing & 12.7 & bath & 28.6\\
364 & 480 & leaping & 14.5 & greetings & 7.5 & 424 & 578 & grilling & 12.6 & meat & 3.8\\
365 & 376 & paddling & 14.5 & large & 22.9 & 425 & 994 & autographing & 12.5 & bitter & 7.7\\
366 & 730 & paddling & 14.4 & camera & 7.2 & 426 & 391 & locking & 12.5 & staircase & 8.0\\
367 & 213 & paddling & 14.4 & red & 16.0 & 427 & 155 & massaging & 12.5 & brown & 18.1\\
368 & 460 & trimming & 14.3 & tomatoes & 15.8 & 428 & 920 & competing & 12.5 & player & 14.6\\
369 & 861 & dusting & 14.3 & swiffer & 10.0 & 429 & 204 & autographing & 12.5 & conference & 3.4\\
370 & 537 & leaping & 14.3 & daughter & 15.4 & 430 & 959 & manicuring & 12.5 & purplish & 13.8\\
371 & 933 & towing & 14.3 & truck & 26.5 & 431 & 896 & bandaging & 12.5 & tape & 4.8\\
372 & 636 & paddling & 14.3 & trees & 15.9 & 432 & 820 & peeling & 12.4 & cutting & 5.6\\
373 & 336 & juggling & 14.2 & fire & 15.2 & 433 & 835 & drumming & 12.4 & circle & 12.5\\
374 & 794 & juggling & 14.2 & boy & 3.2 & 434 & 202 & dropping & 12.4 & surface & 3.7\\
375 & 283 & piloting & 14.1 & statue & 9.4 & 435 & 983 & rinsing & 12.3 & scrubbing & 5.6\\
376 & 20 & singing & 14.1 & camera & 5.8 & 436 & 519 & autographing & 12.3 & camera & 7.4\\
377 & 419 & leaping & 14.0 & flying & 13.2 & 437 & 945 & lecturing & 12.3 & talking & 6.2\\
378 & 507 & racing & 14.0 & track & 10.1 & 438 & 754 & paddling & 12.3 & man & 8.6\\
379 & 445 & driving & 14.0 & cars & 8.8 & 439 & 601 & dropping & 12.2 & coffee & 17.9\\
380 & 11 & crouching & 14.0 & kneeling & 28.1 & 440 & 140 & lecturing & 12.2 & suit & 12.1\\
381 & 74 & autographing & 13.9 & blond & 26.1 & 441 & 87 & fueling & 12.2 & pickup & 8.3\\
382 & 901 & singing & 13.9 & girl & 13.1 & 442 & 408 & paddling & 12.2 & blue & 7.0\\
383 & 313 & leaping & 13.9 & toys & 21.6 & 443 & 333 & draining & 12.2 & coming & 17.4\\
384 & 346 & packing & 13.8 & conveyor & 18.2 & 444 & 979 & lecturing & 12.1 & podium & 8.4\\
385 & 508 & paddling & 13.8 & person & 15.3 & 445 & 871 & falling & 12.1 & waterfall & 10.7\\
386 & 267 & saluting & 13.8 & soldiers & 12.8 & 446 & 53 & paddling & 12.1 & seen & 7.5\\
387 & 452 & drumming & 13.8 & stage & 18.0 & 447 & 32 & paddling & 12.1 & jeans & 1.7\\
388 & 944 & massaging & 13.8 & back & 6.0 & 448 & 488 & pedaling & 12.1 & bike & 5.3\\
389 & 595 & juggling & 13.8 & throws & 6.3 & 449 & 839 & pushing & 12.1 & pushing & 20.8\\
390 & 224 & paddling & 13.8 & day & 4.3 & 450 & 378 & dunking & 12.1 & court & 3.7\\
391 & 619 & shredding & 13.8 & machinery & 7.0 & 451 & 489 & applauding & 12.1 & crowd & 4.0\\
392 & 512 & juggling & 13.7 & t-shirt & 7.2 & 452 & 999 & leaping & 12.1 & children & 4.6\\
393 & 160 & autographing & 13.7 & paper & 11.7 & 453 & 339 & skating & 12.1 & \tiny{skateboarding} & 22.0\\
394 & 390 & pouring & 13.7 & liquid & 21.1 & 454 & 653 & dropping & 12.1 & slow & 6.0\\
395 & 394 & paddling & 13.7 & car & 5.3 & 455 & 225 & autographing & 12.1 & city & 6.3\\
396 & 541 & flossing & 13.6 & fancy & 8.9 & 456 & 30 & leaping & 12.0 & dog & 19.8\\
397 & 396 & massaging & 13.6 & electronical & 6.6 & 457 & 654 & applauding & 12.0 & old & 12.1\\
398 & 321 & standing & 13.4 & performing & 9.8 & 458 & 102 & tattooing & 12.0 & drawing & 10.9\\
399 & 432 & weeding & 13.4 & garden & 16.3 & 459 & 662 & autographing & 12.0 & older & 14.0\\
400 & 71 & bulldozing & 13.3 & tractor & 13.7 & 460 & 219 & talking & 12.0 & turned & 5.5\\
401 & 596 & drenching & 13.3 & window & 30.0 & 461 & 99 & dropping & 12.0 & cartoon & 13.3\\
402 & 177 & autographing & 13.3 & broadcast & 2.3 & 462 & 669 & shaving & 11.9 & legs & 11.4\\
403 & 10 & dialing & 13.3 & jack & 13.6 & 463 & 962 & dropping & 11.9 & winds & 7.4\\
404 & 527 & autographing & 13.3 & street & 5.2 & 464 & 205 & sleeping & 11.9 & child & 10.5\\
405 & 837 & drenching & 13.3 & rain & 9.6 & 465 & 936 & dropping & 11.9 & image & 15.2\\
406 & 293 & leaping & 13.3 & fly & 5.8 & 466 & 728 & applauding & 11.9 & rally & 12.5\\
407 & 867 & dropping & 13.3 & bunch & 3.7 & 467 & 804 & leaping & 11.9 & field & 12.0\\
408 & 426 & weeding & 13.3 & gardening & 11.1 & 468 & 529 & leaping & 11.9 & dog & 4.9\\
409 & 280 & leaping & 13.3 & dog & 20.5 & 469 & 971 & hitchhiking & 11.8 & road & 19.4\\
410 & 331 & autographing & 13.1 & contract & 4.4 & 470 & 666 & applauding & 11.8 & smiling & 13.7\\
411 & 523 & leaping & 13.1 & dancing & 8.8 & 471 & 797 & applauding & 11.7 & \tiny{black-n-white} & 11.6\\
412 & 741 & singing & 13.0 & microphone & 12.5 & 472 & 425 & drumming & 11.7 & filming & 3.1\\
413 & 744 & barbecuing & 13.0 & chef & 19.5 & 473 & 663 & peeling & 11.6 & waist & 5.3\\
414 & 724 & sawing & 13.0 & tree & 5.6 & 474 & 471 & applauding & 11.6 & hands & 10.8\\
415 & 277 & juggling & 13.0 & motion & 3.9 & 475 & 711 & leaping & 11.5 & children & 7.8\\
416 & 16 & dialing & 12.9 & device & 4.9 & 476 & 611 & sleeping & 11.5 & dog & 8.1\\
417 & 984 & destroying & 12.9 & tower & 11.4 & 477 & 715 & paddling & 11.5 & blue & 7.8\\
418 & 917 & dragging & 12.9 & pulling & 20.1 & 478 & 36 & singing & 11.5 & microphone & 24.7\\
419 & 729 & leaping & 12.8 & running & 7.2 & 479 & 700 & tattooing & 11.4 & someone's & 3.2\\
420 & 365 & autographing & 12.8 & walk & 7.5 & 480 & 1017 & applauding & 11.4 & standing & 6.1\\
    \bottomrule
  \end{tabular}
\endgroup
\end{table*}
\setcounter{table}{5}
\begin{table*}
\centering
\begingroup
\setlength{\tabcolsep}{5pt}
\renewcommand{\arraystretch}{1.2}
\scriptsize
  \caption{continued}
  \centering
  \begin{tabular}{c  c  c c  c c  C  c  c c  c c }
    \toprule
         \multirow{3}{*}{Rank} &\multirow{3}{*}{Code}  &  \multicolumn{2}{c}{ Visual Action }  & \multicolumn{2}{c}{ Spoken word } &  \multirow{3}{*}{Rank} &\multirow{3}{*}{Code}  &  \multicolumn{2}{c}{ Visual Action }  & \multicolumn{2}{c}{ Spoken word }\\
      & &\multicolumn{2}{c}{ Top Hypothesis }  &  \multicolumn{2}{c}{ Top Hypothesis } & & &\multicolumn{2}{c}{ Top Hypothesis }  &  \multicolumn{2}{c}{ Top Hypothesis }\\
     & &label & Prc.  & word & F1 && &label & Prc.  & word & F1  \\
    \cline{1-6}\cline{7-12}
481 & 887 & autographing & 11.4 & sidewalk & 4.7 & 541 & 982 & mopping & 9.8 & floor & 16.7\\
482 & 829 & leaping & 11.4 & cat & 14.5 & 542 & 115 & autographing & 9.7 & yelling & 7.6\\
483 & 162 & lecturing & 11.4 & speaking & 7.3 & 543 & 179 & sleeping & 9.7 & tiger & 13.2\\
484 & 852 & swimming & 11.3 & pool & 15.4 & 544 & 877 & autographing & 9.7 & at & 1.6\\
485 & 353 & paddling & 11.2 & trickling & 2.1 & 545 & 84 & paddling & 9.7 & going & 7.8\\
486 & 998 & paddling & 11.2 & green & 22.4 & 546 & 132 & autographing & 9.7 & people & 2.7\\
487 & 565 & manicuring & 11.2 & painting & 8.6 & 547 & 902 & autographing & 9.7 & another & 2.6\\
488 & 129 & drumming & 11.2 & night & 25.9 & 548 & 383 & paddling & 9.6 & buildings & 4.8\\
489 & 21 & paddling & 11.2 & going & 7.6 & 549 & 124 & drumming & 9.6 & silhouette & 3.7\\
490 & 690 & autographing & 11.2 & blond & 7.2 & 550 & 986 & paddling & 9.6 & sliding & 14.0\\
491 & 214 & slipping & 11.1 & snow & 11.0 & 551 & 253 & applauding & 9.5 & tennis & 5.7\\
492 & 454 & paddling & 11.1 & bridge & 5.6 & 552 & 761 & autographing & 9.5 & ground & 5.4\\
493 & 320 & unpacking & 11.1 & boxes & 18.2 & 553 & 925 & rafting & 9.5 & group & 13.9\\
494 & 261 & paddling & 11.1 & down & 3.4 & 554 & 90 & drumming & 9.4 & wearing & 10.2\\
495 & 864 & sleeping & 11.1 & father & 7.1 & 555 & 583 & autographing & 9.4 & standing & 7.9\\
496 & 411 & burying & 11.0 & hole & 16.0 & 556 & 725 & hammering & 9.4 & blacksmith & 11.1\\
497 & 127 & competing & 11.0 & field & 5.2 & 557 & 976 & peeling & 9.4 & closing & 12.9\\
498 & 580 & child+singing & 10.8 & girl & 10.7 & 558 & 922 & drenching & 9.4 & driving & 16.8\\
499 & 849 & paddling & 10.8 & slowly & 8.4 & 559 & 198 & singing & 9.4 & tide & 9.1\\
500 & 285 & autographing & 10.7 & dress & 8.1 & 560 & 144 & screwing & 9.4 & machine & 7.3\\
501 & 721 & autographing & 10.7 & middle-aged & 2.6 & 561 & 607 & extinguishing & 9.4 & spraying & 22.7\\
502 & 88 & leaping & 10.7 & wall & 10.1 & 562 & 195 & racing & 9.4 & cars & 23.0\\
503 & 769 & autographing & 10.6 & table & 10.2 & 563 & 913 & drumming & 9.3 & sitting & 6.6\\
504 & 91 & autographing & 10.6 & she & 8.1 & 564 & 366 & bulldozing & 9.3 & trainer & 2.6\\
505 & 119 & jumping & 10.6 & rope & 8.8 & 565 & 703 & leaping & 9.3 & cats & 4.3\\
506 & 448 & paddling & 10.6 & hat & 8.3 & 566 & 367 & autographing & 9.3 & holding & 6.8\\
507 & 831 & skating & 10.5 & park & 20.6 & 567 & 377 & autographing & 9.3 & hallway & 11.8\\
508 & 906 & leaping & 10.5 & store & 9.6 & 568 & 173 & raining & 9.2 & cartoon & 25.4\\
509 & 344 & discussing & 10.5 & restaurant & 25.7 & 569 & 86 & competing & 9.2 & field & 13.1\\
510 & 847 & cheering & 10.5 & competition & 4.3 & 570 & 328 & autographing & 9.2 & walking & 13.2\\
511 & 357 & shaving & 10.4 & his & 5.3 & 571 & 258 & leaping & 9.2 & kids & 7.1\\
512 & 904 & running & 10.4 & running & 13.1 & 572 & 487 & autographing & 9.2 & giving & 1.5\\
513 & 193 & paddling & 10.4 & someone & 16.2 & 573 & 385 & ironing & 9.2 & clothes & 15.8\\
514 & 192 & applauding & 10.3 & motocross & 6.2 & 574 & 598 & raining & 9.1 & cartoon & 17.6\\
515 & 230 & autographing & 10.3 & looking & 10.8 & 575 & 128 & surfing & 9.1 & standstill & 9.8\\
516 & 534 & sleeping & 10.3 & bag & 4.5 & 576 & 851 & lecturing & 9.1 & upside & 12.5\\
517 & 550 & peeling & 10.2 & bowl & 15.7 & 577 & 649 & pouring & 9.1 & concrete & 11.1\\
518 & 159 & autographing & 10.2 & ward & 4.3 & 578 & 695 & sleeping & 9.1 & couch & 12.8\\
519 & 314 & leaping & 10.2 & mixed-race & 4.0 & 579 & 70 & autographing & 9.1 & people & 4.4\\
520 & 709 & leaping & 10.1 & animals & 4.8 & 580 & 197 & yawning & 9.1 & couch & 30.6\\
521 & 95 & sprinkling & 10.1 & sprinkler & 10.8 & 581 & 446 & applauding & 9.0 & many & 8.8\\
522 & 227 & sleeping & 10.1 & oh & 2.7 & 582 & 351 & singing & 9.0 & bright & 3.2\\
523 & 935 & applauding & 10.1 & perch & 12.5 & 583 & 287 & paddling & 9.0 & bird & 15.2\\
524 & 176 & typing & 10.1 & office & 5.3 & 584 & 821 & drumming & 9.0 & kayakers & 3.8\\
525 & 1011 & drumming & 10.0 & boy & 17.0 & 585 & 310 & applauding & 9.0 & smiling & 12.9\\
526 & 683 & competing & 10.0 & game & 7.1 & 586 & 203 & paddling & 8.9 & video & 3.7\\
527 & 185 & knitting & 10.0 & stitching & 8.2 & 587 & 624 & crushing & 8.9 & greenfield & 20.3\\
528 & 289 & dropping & 10.0 & ground & 16.5 & 588 & 696 & autographing & 8.9 & man & 9.7\\
529 & 899 & reaching & 10.0 & church & 20.8 & 589 & 514 & paddling & 8.9 & behind & 2.8\\
530 & 767 & playing & 10.0 & overwatch & 6.7 & 590 & 886 & falling & 8.9 & shine & 5.6\\
531 & 796 & paddling & 10.0 & base & 3.7 & 591 & 451 & peeling & 8.9 & carrots & 5.2\\
532 & 161 & discussing & 9.9 & family & 10.8 & 592 & 953 & autographing & 8.8 & outside & 12.5\\
533 & 782 & leaping & 9.9 & doing & 12.9 & 593 & 975 & paddling & 8.8 & building & 1.8\\
534 & 850 & autographing & 9.9 & american & 1.9 & 594 & 643 & carving & 8.8 & working & 12.4\\
535 & 620 & leaping & 9.8 & bridge & 4.1 & 595 & 418 & autographing & 8.8 & suit & 13.1\\
536 & 992 & leaping & 9.8 & point-of-view & 6.9 & 596 & 481 & autographing & 8.7 & woman & 9.1\\
537 & 547 & grilling & 9.8 & crawling & 17.3 & 597 & 756 & paddling & 8.7 & wearing & 3.2\\
538 & 891 & paddling & 9.8 & on & 2.5 & 598 & 670 & signing & 8.7 & table & 7.2\\
539 & 340 & dusting & 9.8 & clean. & 5.2 & 599 & 785 & autographing & 8.7 & standing & 2.1\\
540 & 659 & storming & 9.8 & yard & 18.2 & 600 & 787 & drumming & 8.7 & sitting & 7.5\\
    \bottomrule
  \end{tabular}
\endgroup
\end{table*}
\setcounter{table}{5}
\begin{table*}
\centering
\begingroup
\setlength{\tabcolsep}{5pt}
\renewcommand{\arraystretch}{1.2}
\scriptsize
  \caption{continued}
  \centering
  \begin{tabular}{c  c  c c  c c  C  c  c c  c c }
    \toprule
         \multirow{3}{*}{Rank} &\multirow{3}{*}{Code}  &  \multicolumn{2}{c}{ Visual Action }  & \multicolumn{2}{c}{ Spoken word } &  \multirow{3}{*}{Rank} &\multirow{3}{*}{Code}  &  \multicolumn{2}{c}{ Visual Action }  & \multicolumn{2}{c}{ Spoken word }\\
      & &\multicolumn{2}{c}{ Top Hypothesis }  &  \multicolumn{2}{c}{ Top Hypothesis } & & &\multicolumn{2}{c}{ Top Hypothesis }  &  \multicolumn{2}{c}{ Top Hypothesis }\\
     & &label & Prc.  & word & F1 && &label & Prc.  & word & F1  \\
    \cline{1-6}\cline{7-12}
601 & 723 & autographing & 8.7 & something & 6.1 & 661 & 980 & kicking & 7.7 & shooting & 12.5\\
602 & 719 & talking & 8.7 & toddler & 17.9 & 662 & 238 & sleeping & 7.7 & squirrel & 12.8\\
603 & 209 & autographing & 8.7 & hair & 8.4 & 663 & 360 & injecting & 7.7 & person & 2.9\\
604 & 521 & rafting & 8.7 & people & 8.4 & 664 & 853 & camping & 7.7 & tent & 8.9\\
605 & 98 & applauding & 8.6 & stand & 6.2 & 665 & 652 & autographing & 7.6 & single & 1.5\\
606 & 969 & leaping & 8.6 & kids & 7.2 & 666 & 893 & watering & 7.6 & watering & 8.4\\
607 & 770 & flossing & 8.6 & explaining & 7.5 & 667 & 546 & piloting & 7.6 & lyrics & 3.4\\
608 & 616 & leaping & 8.5 & snow & 20.0 & 668 & 674 & applauding & 7.6 & night & 8.4\\
609 & 410 & erupting & 8.5 & explodes & 7.5 & 669 & 881 & autographing & 7.5 & table & 7.6\\
610 & 12 & paddling & 8.5 & distance & 13.6 & 670 & 236 & hammering & 7.5 & wooden & 9.5\\
611 & 750 & flossing & 8.5 & drinking & 8.5 & 671 & 778 & leaping & 7.5 & house & 8.7\\
612 & 1009 & autographing & 8.5 & street & 17.1 & 672 & 260 & flossing & 7.5 & smiling & 19.2\\
613 & 463 & slicing & 8.5 & pieces & 11.7 & 673 & 399 & paddling & 7.5 & of & 6.5\\
614 & 843 & autographing & 8.5 & speaking & 4.8 & 674 & 146 & autographing & 7.5 & language & 4.9\\
615 & 772 & paddling & 8.5 & workers & 12.7 & 675 & 745 & autographing & 7.5 & sitting & 10.0\\
616 & 781 & leaping & 8.5 & involving & 7.4 & 676 & 207 & paddling & 7.5 & man & 12.5\\
617 & 757 & flossing & 8.4 & caption & 10.4 & 677 & 732 & smelling & 7.5 & flowers & 36.6\\
618 & 793 & pointing & 8.3 & gameplay & 16.7 & 678 & 647 & autographing & 7.4 & smashes & 3.3\\
619 & 403 & racing & 8.3 & car & 5.3 & 679 & 894 & splashing & 7.4 & plastic & 9.2\\
620 & 988 & clipping & 8.3 & shoe & 8.3 & 680 & 416 & drumming & 7.4 & group & 9.8\\
621 & 502 & paddling & 8.3 & going & 1.5 & 681 & 492 & autographing & 7.4 & fans & 1.4\\
622 & 343 & paddling & 8.3 & over & 2.7 & 682 & 467 & drumming & 7.4 & child & 8.0\\
623 & 450 & shaving & 8.3 & chef & 5.2 & 683 & 573 & wrapping & 7.4 & box & 9.8\\
624 & 765 & paddling & 8.2 & gymnast & 4.8 & 684 & 381 & autographing & 7.4 & he & 3.0\\
625 & 476 & paddling & 8.2 & trees & 8.4 & 685 & 855 & autographing & 7.3 & gentleman & 2.6\\
626 & 818 & autographing & 8.2 & vest & 3.3 & 686 & 939 & peeling & 7.3 & close & 7.0\\
627 & 746 & autographing & 8.2 & street & 13.7 & 687 & 494 & peeling & 7.3 & hands & 4.9\\
628 & 122 & applauding & 8.2 & people & 7.8 & 688 & 575 & paddling & 7.3 & a & 6.0\\
629 & 275 & leaping & 8.2 & workout & 4.2 & 689 & 581 & smashing & 7.3 & building & 21.9\\
630 & 592 & hammering & 8.2 & piles & 3.9 & 690 & 142 & stopping & 7.3 & characters & 11.4\\
631 & 1003 & leaping & 8.2 & around & 5.9 & 691 & 599 & autographing & 7.3 & two & 5.6\\
632 & 7 & paddling & 8.1 & and & 3.2 & 692 & 309 & paddling & 7.2 & shooting & 4.8\\
633 & 257 & raining & 8.1 & blown & 20.7 & 693 & 264 & drumming & 7.2 & bedroom & 3.6\\
634 & 170 & leaping & 8.1 & running & 5.5 & 694 & 919 & autographing & 7.2 & hands & 12.4\\
635 & 341 & flossing & 8.1 & how & 6.1 & 695 & 47 & autographing & 7.2 & woman & 6.4\\
636 & 354 & sewing & 8.1 & machine & 14.2 & 696 & 570 & paddling & 7.1 & we & 2.0\\
637 & 600 & paddling & 8.1 & each & 6.1 & 697 & 965 & paddling & 7.1 & red & 3.4\\
638 & 677 & rolling & 8.1 & cooks & 6.9 & 698 & 361 & autographing & 7.1 & upright & 3.6\\
639 & 133 & sleeping & 8.1 & string & 7.3 & 699 & 934 & peeling & 7.1 & putting & 7.9\\
640 & 678 & leaping & 8.0 & tree & 7.5 & 700 & 594 & sitting & 7.1 & inject & 8.0\\
641 & 466 & injecting & 8.0 & close & 18.2 & 701 & 186 & draining & 7.1 & house & 20.9\\
642 & 1006 & stirring & 8.0 & pot & 5.2 & 702 & 809 & paddling & 7.1 & man & 11.0\\
643 & 212 & crying & 8.0 & helping & 7.0 & 703 & 783 & paddling & 7.1 & with & 1.6\\
644 & 889 & autographing & 8.0 & young & 4.0 & 704 & 151 & autographing & 7.0 & store & 6.5\\
645 & 437 & manicuring & 7.9 & fingers & 8.3 & 705 & 474 & paddling & 7.0 & person & 9.2\\
646 & 468 & jumping & 7.9 & motorcycle & 9.9 & 706 & 1021 & paddling & 7.0 & decorated & 8.9\\
647 & 633 & applauding & 7.9 & show & 7.9 & 707 & 359 & paddling & 7.0 & picture & 5.7\\
648 & 59 & drumming & 7.9 & watching & 14.2 & 708 & 740 & applauding & 6.9 & people & 6.1\\
649 & 814 & peeling & 7.9 & someone & 4.7 & 709 & 614 & autographing & 6.9 & knick-knack & 5.3\\
650 & 441 & leaping & 7.9 & jeans & 14.1 & 710 & 237 & leaping & 6.9 & inflating & 6.7\\
651 & 775 & inflating & 7.9 & chair & 16.5 & 711 & 567 & autographing & 6.9 & and & 2.8\\
652 & 46 & autographing & 7.9 & woman & 23.8 & 712 & 135 & signing & 6.9 & sitting & 6.6\\
653 & 824 & autographing & 7.9 & field & 2.3 & 713 & 506 & drumming & 6.9 & guys & 7.3\\
654 & 968 & dusting & 7.9 & floor & 5.4 & 714 & 28 & autographing & 6.9 & hey & 7.0\\
655 & 544 & reaching & 7.8 & climbing & 13.0 & 715 & 456 & applauding & 6.9 & animated & 4.9\\
656 & 389 & autographing & 7.8 & front & 14.3 & 716 & 768 & autographing & 6.9 & holding & 7.4\\
657 & 819 & lecturing & 7.8 & laughing & 18.1 & 717 & 879 & leaping & 6.9 & exercising & 5.1\\
658 & 498 & shaving & 7.8 & pink & 18.5 & 718 & 628 & drumming & 6.9 & men & 10.4\\
659 & 9 & paddling & 7.7 & green & 5.2 & 719 & 80 & discussing & 6.9 & women & 8.6\\
660 & 927 & singing & 7.7 & saying & 13.9 & 720 & 586 & applauding & 6.8 & haired & 7.7\\
    \bottomrule
  \end{tabular}
\endgroup
\end{table*}
\setcounter{table}{5}
\begin{table*}
\centering
\begingroup
\setlength{\tabcolsep}{5pt}
\renewcommand{\arraystretch}{1.2}
\scriptsize
  \caption{continued}
  \centering
  \begin{tabular}{c  c  c c  c c  C  c  c c  c c }
    \toprule
         \multirow{3}{*}{Rank} &\multirow{3}{*}{Code}  &  \multicolumn{2}{c}{ Visual Action }  & \multicolumn{2}{c}{ Spoken word } &  \multirow{3}{*}{Rank} &\multirow{3}{*}{Code}  &  \multicolumn{2}{c}{ Visual Action }  & \multicolumn{2}{c}{ Spoken word }\\
      & &\multicolumn{2}{c}{ Top Hypothesis }  &  \multicolumn{2}{c}{ Top Hypothesis } & & &\multicolumn{2}{c}{ Top Hypothesis }  &  \multicolumn{2}{c}{ Top Hypothesis }\\
     & &label & Prc.  & word & F1 && &label & Prc.  & word & F1  \\
    \cline{1-6}\cline{7-12}
721 & 1004 & pouring & 6.8 & color & 6.8 & 734 & 167 & peeling & 6.7 & into & 3.5\\
722 & 72 & injecting & 6.8 & mixing & 4.7 & 735 & 428 & leaping & 6.7 & doing & 2.2\\
723 & 876 & autographing & 6.8 & arena & 3.8 & 736 & 300 & paddling & 6.6 & is & 2.3\\
724 & 878 & gambling & 6.8 & game & 9.4 & 737 & 786 & paddling & 6.6 & background & 1.7\\
725 & 114 & paddling & 6.8 & man & 15.7 & 738 & 272 & leaping & 6.6 & truck & 8.5\\
726 & 764 & raising & 6.8 & trash & 10.0 & 739 & 297 & applauding & 6.6 & rustic & 4.5\\
727 & 326 & leaping & 6.8 & zooming & 3.1 & 740 & 303 & paddling & 6.6 & down & 1.0\\
728 & 217 & surfing & 6.8 & surfer & 2.0 & 741 & 93 & paddling & 6.5 & car & 3.3\\
729 & 107 & pouring & 6.8 & bottle & 5.8 & 742 & 542 & baking & 6.5 & cupcake & 9.2\\
730 & 749 & injecting & 6.7 & person's & 2.2 & 743 & 733 & discussing & 6.5 & men & 7.3\\
731 & 612 & autographing & 6.7 & disgust & 4.8 & 744 & 23 & leaping & 6.5 & cross & 5.5\\
732 & 834 & autographing & 6.7 & outside & 12.7 & 745 & 104 & injecting & 6.5 & beard & 2.6\\
733 & 655 & marrying & 6.7 & couple & 7.5 & 746 & 279 & leaping & 6.5 & garden & 4.4\\
    \bottomrule
  \end{tabular}
\endgroup
\end{table*}

\end{document}